\documentclass[10pt, logo, twocolumn, copyright]{nvidiatechreport}
\usepackage{mdframed}
\usepackage[utf8]{inputenc}
\usepackage[T1]{fontenc}
\usepackage{amsfonts}
\usepackage{nicefrac}
\usepackage{microtype}
\usepackage[dvipsnames]{xcolor}
\usepackage{multirow}
\usepackage{multicol}
\usepackage{graphicx}
\usepackage[numbers]{natbib}
\usepackage{tabto}
\usepackage{xspace}
\usepackage{amsmath}
\usepackage{adjustbox}
\usepackage{enumitem}
\usepackage{wrapfig}
\usepackage{dblfloatfix}
\usepackage{float}
\usepackage{booktabs}
\usepackage{tabularx}
\usepackage{arydshln}
\usepackage{colortbl}
\usepackage{makecell}
\usepackage{hhline}
\usepackage{array}
\usepackage{diagbox}
\usepackage{fp}
\usepackage{hyperref}
\usepackage{url}

\newcolumntype{L}[1]{>{\raggedright\let\newline\\\arraybackslash\hspace{0pt}}m{#1}}
\newcolumntype{C}[1]{>{\centering\let\newline\\\arraybackslash\hspace{0pt}}m{#1}}
\newcolumntype{R}[1]{>{\raggedleft\let\newline\\\arraybackslash\hspace{0pt}}m{#1}}

\newcommand{\fig}[1]{Figure~\ref{#1}}

\newcommand{\ignorethis}[1]{}

\makeatletter
\DeclareRobustCommand\onedot{\futurelet\@let@token\@onedot}
\def\@onedot{\ifx\@let@token.\else.\null\fi\xspace}

\makeatother

\makeatletter
\def\adl@drawiv#1#2#3{%
        \hskip.5\tabcolsep
        \xleaders#3{#2.5\@tempdimb #1{1}#2.5\@tempdimb}%
                #2\z@ plus1fil minus1fil\relax
        \hskip.5\tabcolsep}
\newcommand{\cdashlinelr}[1]{%
  \noalign{\vskip\aboverulesep
           \global\let\@dashdrawstore\adl@draw
           \global\let\adl@draw\adl@drawiv}
  \cdashline{#1}
  \noalign{\global\let\adl@draw\@dashdrawstore
           \vskip\belowrulesep}}
\makeatother

\definecolor{citecolor}{HTML}{0071bc}
\definecolor{mydarkblue}{rgb}{0,0.08,1}
\definecolor{mydarkgreen}{rgb}{0.02,0.6,0.02}
\definecolor{mydarkred}{rgb}{0.8,0.02,0.02}
\definecolor{mydarkorange}{rgb}{0.40,0.2,0.02}
\definecolor{mypurple}{RGB}{111,0,255}
\definecolor{myred}{rgb}{1.0,0.0,0.0}
\definecolor{mygold}{rgb}{0.75,0.6,0.12}
\definecolor{mydarkgray}{rgb}{0.66, 0.66, 0.66}

\definecolor{darkblue}{rgb}{0,0.08,1}
\definecolor{darkgreen}{rgb}{0.02,0.6,0.02}
\definecolor{darkred}{rgb}{0.8,0.02,0.02}
\definecolor{darkorange}{rgb}{0.40,0.2,0.02}
\definecolor{darkpurple}{RGB}{111,0,255}

\definecolor{spc}{RGB}{119, 107, 170}
\definecolor{pct}{rgb}{0.7, 0, 0.2}

\newcommand\rt{ \rowcolor{teal!15}}

\newcommand\rc{ \rowcolor{cyan!15}}

\newcommand{\myparagraph}[1]{\vspace{0pt}\paragraph{#1}}

\definecolor{mydarkblue}{rgb}{0,0.08,1}

\def\method{STORM\xspace}

\usepackage{subcaption}
\usepackage[flushmargin]{footmisc}

\title{STORM: Token-Efficient Long Video Understanding for Multimodal LLMs}

\begin{document}

\author{Jindong Jiang\textsuperscript{1,2,$\dag$,$\star$} \quad
Xiuyu Li\textsuperscript{1,3,$\dag$,$\star$} \quad
Zhijian Liu\textsuperscript{1} \quad
Muyang Li\textsuperscript{1,4,$\star$} \quad
Guo Chen\textsuperscript{1,5,$\star$} \quad
Zhiqi Li\textsuperscript{1,5,$\star$} \quad
De-An Huang\textsuperscript{1} \quad
Guilin Liu\textsuperscript{1} \quad
Zhiding Yu\textsuperscript{1} \quad
Kurt Keutzer\textsuperscript{3} \quad
Sungjin Ahn\textsuperscript{6} \quad
Jan Kautz\textsuperscript{1} \quad
Hongxu Yin\textsuperscript{1} \quad
Yao Lu\textsuperscript{1} \quad
Song Han\textsuperscript{1,4} \quad
Wonmin Byeon\textsuperscript{1} \\~\\
\textsuperscript{1}NVIDIA \quad \textsuperscript{2}Rutgers University \quad \textsuperscript{3}UC Berkeley \quad \textsuperscript{4}MIT \quad \textsuperscript{5}Nanjing University \quad \textsuperscript{6}KAIST \\
\textsuperscript{$\dag$}Equal contribution \quad \textsuperscript{$\star$}Work performed during internship at NVIDIA}

\begin{abstract}
Recent advances in video-based multimodal large language models (Video-LLMs) have significantly improved video understanding by processing videos as sequences of image frames. However, many existing methods treat frames independently in the vision backbone, lacking explicit temporal modeling, which limits their ability to capture dynamic patterns and efficiently handle long videos. To address these limitations, we introduce \method (\textcolor{blue}{\textbf{S}}patiotemporal \textcolor{blue}{\textbf{TO}}ken \textcolor{blue}{\textbf{R}}eduction for \textcolor{blue}{\textbf{M}}ultimodal LLMs), a novel architecture incorporating a dedicated temporal encoder between the image encoder and the LLM.  Our temporal encoder leverages the Mamba State Space Model to integrate temporal information into image tokens, generating enriched representations that preserve inter-frame dynamics across the entire video sequence. 
This enriched encoding not only enhances video reasoning capabilities but also enables effective token reduction strategies, including test-time sampling and training-based temporal and spatial pooling, substantially reducing computational demands on the LLM without sacrificing key temporal information.
By integrating these techniques, our approach simultaneously reduces training and inference latency while improving performance, enabling efficient and robust video understanding over extended temporal contexts. Extensive evaluations show that \method achieves state-of-the-art results across various long video understanding benchmarks (more than 5\% improvement on MLVU and LongVideoBench) while reducing the computation costs by up to $8\times$ and the decoding latency by 2.4-2.9$\times$ for the fixed numbers of input frames. Project website is available at \href{https://research.nvidia.com/labs/lpr/storm}{here}.
\end{abstract}

\maketitle

\section{Introduction}

\begin{figure*}[!t]
    \centering
    \includegraphics[width=\textwidth]{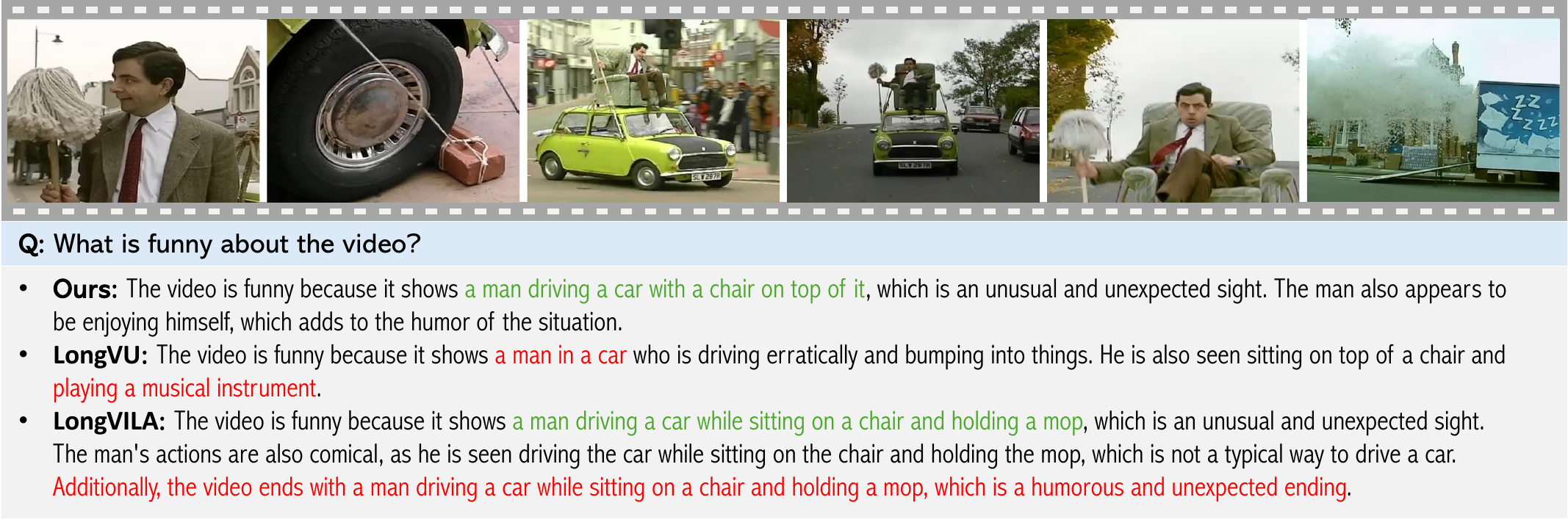}
    \caption{ \textbf{Open-Ended Video Understanding.} We show \method's ability to handle free-form queries about complex long video scenes. By employing the Mamba-based temporal encoder to capture essential spatiotemporal cues while compressing redundant frame information, \method enables efficient, accurate long-video understanding and outperforms existing methods on a wide range of video understanding tasks.}
    \label{fig:teaser_example}
\end{figure*}

Recent advancements in video-based multimodal Large Language Models (Video-LLMs) have improved AI systems' ability to understand video content~\citep{liu2024visual, lin2024vila, li2023videochat, damonlpsg2024videollama2, wang2022internvideo, wang2024internvideo2}. These models typically process videos as sequences of individual frames, encoding each frame independently before feeding representations to a large language model (LLM) for temporal reasoning~\citep{li2023videochat, damonlpsg2024videollama2, lin2024vila}. 
Despite leveraging powerful image encoders and LLMs~\citep{radford2021learning, zhai2023sigmoid}, these approaches exhibit fundamental limitations in video processing, particularly for long videos.  The lack of explicit temporal encoding in image tokens forces LLMs to infer temporal relationships from static image sequences, creating substantial computational burden. Moreover, to manage computational overhead, existing methods like LongVILA~\citep{longvila} and LongVA~\citep{zhang2024longva}
often use naive frame subsampling to reduce LLM processing tokens, resulting in significant information loss, potentially eliminating critical details necessary for comprehensive video understanding. 
Recently, LongVU~\citep{shen2024longvu} proposes token compression by adaptively selecting visual tokens based on user queries. However, unlike our approach, LongVU lacks a dedicated module for propagating spatiotemporal information before token selection, potentially limiting the model's ability to preserve critical information after token reduction.

In this paper, we propose \method ({\textbf{S}}patiotemporal {\textbf{TO}}ken {\textbf{R}}eduction for {\textbf{M}}ultimodal LLMs), which introduces a novel temporal encoder between the image encoder and the LLM. This design integrates temporal dynamics earlier in the model pipeline, significantly enhancing the temporal reasoning capabilities of Video-LLMs while enabling substantial downstream token reduction. We employ Mamba ~\citep{gu2024mamba} as the core of our temporal layer to allow both computation efficiency and generalization to extended temporal contexts. 
While there are existing Mamba-based video understanding models~\citep{li2024videomamba, park2024videomamba, lu2024videomambapro, chen2024video, li2024mamba}, unlike ours, these methods do not integrate Mamba with LLMs for multimodal tasks, and importantly, they mainly focus on replacing traditional backbones with Mamba architectures without explicitly leveraging Mamba's ability for reducing video redundancy and enabling visual token compression.  

The key advantage of the Mamba layer is its ability to compress historical information into state representations. As consecutive frames in the video input often contain redundant information, our temporal encoder efficiently processes and propagates temporal information across the entire video sequence. The resulting visual tokens inherently encapsulate temporal history, allowing us to extract fewer tokens for LLM processing while preserving key information. We explore both training-free token subsampling and training-based compression methods (temporal and spatial token compression). Unlike previous frame-subsampling approaches, our method preserves essential temporal information in a compressed format, reducing computational load while improving performance through more comprehensive representation in a compact token space.
Empirical evaluations demonstrate that our method significantly outperform baselines, especially on long-video inputs, while simultaneously achieving strong computational efficiency.

\section{Related Work} \label{sec:related_work}
\myparagraph{Vision Language Model Architecture}

Vision language models (VLMs) primarily adopt two paradigms to process visual input. The first paradigm freezes language model weights and integrates visual information via cross-attention mechanisms~\cite{alayrac2022flamingo}, while the second paradigm utilizes a pre-trained image encoder, such as CLIP~\cite{radford2021learning} or SigLIP~\cite{zhai2023sigmoid}, to convert images into tokens. These tokens are then concatenated with text tokens and input into the language model~\cite{lin2024vila,liu2024visual,driess2023palm}. This approach can be naturally extended to video understanding by treating videos as sequences of images processed by the vision encoder~\citep{li2023videochat, damonlpsg2024videollama2}. To enhance video processing, some works introduce specialized video encoders. For instance, InternVideo~\cite{wang2022internvideo, wang2024internvideo2} uses VideoMAE~\cite{tong2022videomae} as a video encoder, while Kangaroo~\cite{liu2024kangaroo} integrates depth-wise 3D convolution for fusing video tokens. In this work, we retain SigLIP as the vision encoder and focus on enhancing long video understanding by incorporating a linear-complexity temporal module in the Mamba~\citep{gu2024mamba} architecture. Positioned between the SigLIP vision encoder and the language model, this module efficiently improves spatial-temporal modeling effectiveness.

\myparagraph{Long Video Understanding}
Understanding long videos with VLMs presents significant challenges in both accuracy and efficiency. Previous approaches have employed long-context language models trained on short-context video data to enable long video comprehension~\citep{zhang2024longva}. However, these methods lack sufficient long video training data and incur high computational costs during both training and inference as the number of frames increases. LongVILA~\citep{longvila} addresses these challenges through a multi-modal sequence parallelism system that directly handles long video data during training and inference, but this approach requires customized system implementations tailored for multi-GPU setups. Another line of research focuses on token reduction to shorten input sequences, thereby enabling efficient inference for long videos~\citep{li2023llamavid, weng2025longvlm, ye2024voco, ryoo2024xgenmmvid, shu2024videoxl, shen2024longvu}. For instance, VoCO-LLaMA~\citep{ye2024voco} and VideoXL~\citep{shu2024videoxl} use recursive KV cache compression learnt in an end-to-end manner, and LongVU~\citep{shen2024longvu} leverages DINO features for frame selection and inter-frame similarity to reduce tokens. Despite these diverse strategies, direct pooling along the temporal or spatial dimensions often performs sufficiently well, with additional gains being marginal. In this paper, we apply temporal and spatial pooling for token reduction, achieving superior performance when combined with our temporal projector.

\myparagraph{Mamba for Video Understanding}
Recent advances in linear state space models such as Mamba~\citep{gu2024mamba, dao2024transformers} have sparked extensive exploration in applying them to video understanding tasks. Due to their sub-quadratic computation complexity, Mamba models achieve significant efficiency improvements compared to transformer-based architectures while still delivering competitive performance~\citep{gu2024mamba, dao2024transformers, waleffe2024empirical, wang2024mamba, zuo2024falcon}. These properties make Mamba particularly suitable for video processing as the models are required to process long sequence inputs. For example, VideoMamba in~\citep{li2024videomamba} and VideoMamba (identical model naming) in~\citep{park2024videomamba} use Mamba-based visual backbone in video models and demonstrates the model's strong ability to capture both local redundancy and long-term spatiotemporal dependencies. 
VideoMambaPro~\citep{lu2024videomambapro} proposes to improve Mamba's video understanding ability by applying masking and residual connection during the backward scan.
The Video Mamba Suite~\citep{chen2024video} further explores various architectures to integrate Mamba into existing video understanding models, demonstrating favorable efficiency-performance trade-offs for long sequence inputs. Mamba-ND~\citep{li2024mamba} aims to improve Mamba's performance on multi-dimensional data by investigating design choices such as SSM layer structure and scanning order within and across dimensions. However, unlike our approach, these works do not directly apply Mamba for Multimodal LLMs. More importantly, they primarily focus on replacing traditional backbones with Mamba architectures without explicitly leveraging Mamba's unique ability to summarize historical information for reducing video redundancy and enabling visual token compression. Our paper addresses this research gap by proposing STORM, which proves to be both effective and efficient for video understanding while significantly reducing computational demands.

\myparagraph{Concurrent Work} Recently, BIMBA~\citep{islam2025bimba} explores a similar architecture for long-video understanding, reporting similar benefits through empirical evaluation. We are encouraged to see these independent findings further support the hypothesis.

\section{Method}

\begin{figure*}[!t]
    \centering
    \includegraphics[width=\textwidth]{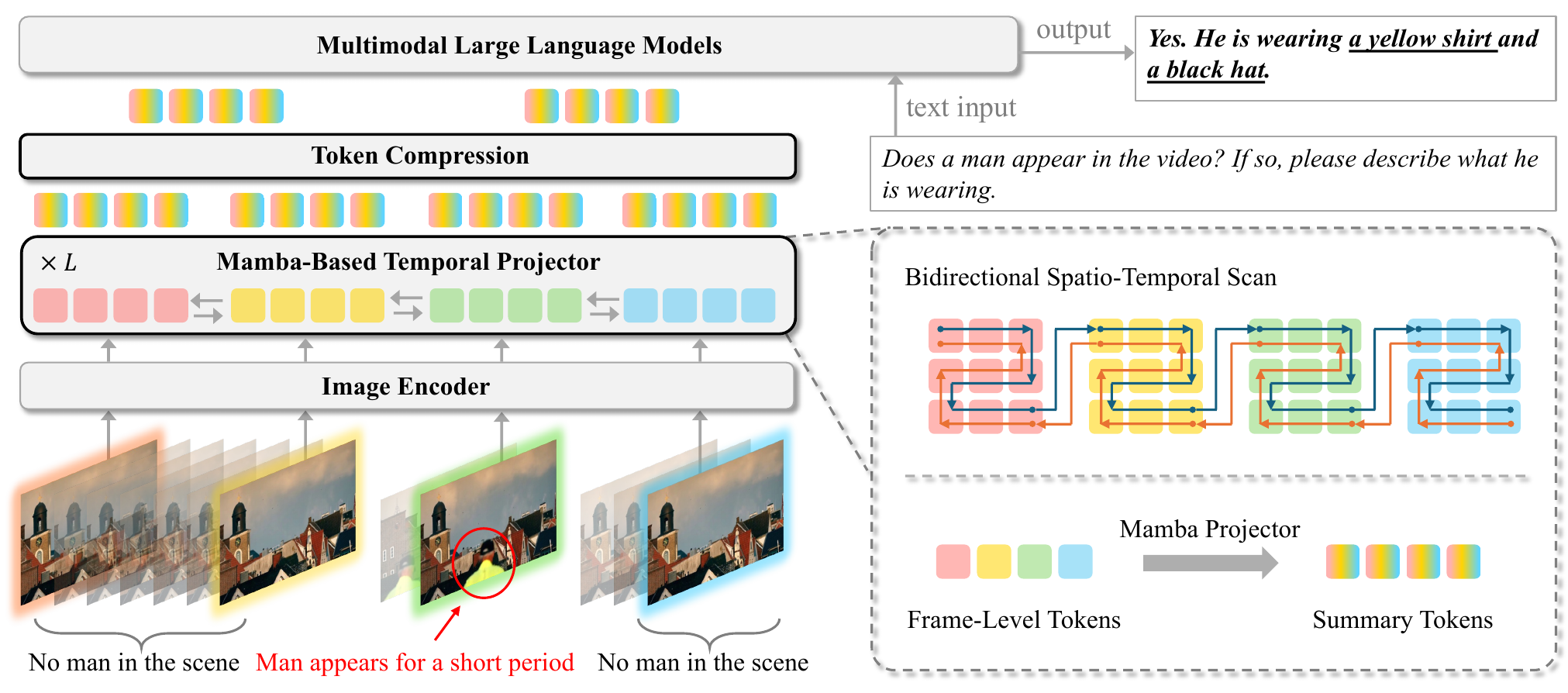}
    \caption{\textbf{Overview of \method pipeline.} \method integrates a Mamba-based temporal projector between the image encoder and LLM. This projector performs spatiotemporal scanning to embed temporal information directly into visual tokens. The resulting Summary Tokens encapsulate temporal history, enabling efficient downstream token reduction while preserving essential video dynamics.
    }
    \label{fig:overview}
\end{figure*}

This section introduces the Mamba-based temporal projector architecture and explores several token compression techniques for efficient long video processing. We investigate both training-based and training-free token compression strategies across temporal and spatial dimensions. We begin with an overview of our Mamba-based temporal projector, followed by a discussion of the token compression methods. The overview of our method is shown in \autoref{fig:overview}.

\subsection{Preliminaries}

\textbf{State Space Models (SSMs)} A State Space Model (SSM)~\citep{fu2023h3, poli2023hyena, gu2024mamba, mamba2} establishes a linear transformation between an input sequence $\boldsymbol{x}_{1:T} \in \mathbb{R}^{T \times D}$ and an output sequence $\boldsymbol{y}_{1:T} \in \mathbb{R}^{T \times D}$ through the following recurrent process:
\begin{equation}
    \begin{aligned}
        \boldsymbol{h}_t &= \boldsymbol{\overline{A}}_t \boldsymbol{h}_{t-1} + \boldsymbol{\overline{B}}_t \boldsymbol{x}_t\ , \\
        \boldsymbol{y}_t &= \boldsymbol{C}_t \boldsymbol{h}_t\ .
    \end{aligned}
\end{equation}

Here, $T$ is the sequence length; $\boldsymbol{x}_t, \boldsymbol{y}_t \in \mathbb{R}^{D}$ are input and output vectors at time $t$; and $\boldsymbol{h}_t \in \mathbb{R}^{H}$ is the hidden state summarizing the history $\boldsymbol{x}_{\leq t}$. The matrices $\boldsymbol{\overline{A}}_t \in \mathbb{R}^{H \times H}$, $\boldsymbol{\overline{B}}_t \in \mathbb{R}^{H \times D}$, and $\boldsymbol{C}_t \in \mathbb{R}^{D \times H}$ are parameterized with learnable weights designed to facilitate the modeling of long-term dependencies. When $\boldsymbol{\overline{A}}_t, \boldsymbol{\overline{B}}_t, \boldsymbol{C}_t$ are time-invariant (constant over $t$), the computation of $\boldsymbol{y}_{1:T}$ can be parallelized, enabling efficient training and inference.

\noindent \textbf{Mamba} Recently, Mamba \cite{gu2024mamba} propose to condition these matrices on the input $\boldsymbol{x}_t$ to enhance the sequence modeling capabilities of SSMs. Specifically, Mamba employs learnable functions $\boldsymbol{\overline{A}} \colon \mathbb{R}^{D} \to \mathbb{R}^{H \times H}$, $\boldsymbol{\overline{B}} \colon \mathbb{R}^{D} \to \mathbb{R}^{H \times D}$, and $\boldsymbol{C} \colon \mathbb{R}^{D} \to \mathbb{R}^{D \times H}$ to generate input-dependent matrices as follows:
\begin{align}
    \boldsymbol{\overline{A}}_t &= \boldsymbol{\overline{A}}(\boldsymbol{x}_t)\ , \quad \boldsymbol{\overline{B}}_t = \boldsymbol{\overline{B}}(\boldsymbol{x}_t)\ , \quad \boldsymbol{C}_t = \boldsymbol{C}(\boldsymbol{x}_t)\ .
\end{align}
This approach allows the model to dynamically emphasize or suppress information based on the current input, thereby enabling more flexible and adaptive sequence modeling. Additionally, Mamba leverages a hardware-aware parallel algorithm to ensuring that the input-dependent matrices does not hinder the training and inference efficiency inherent to SSMs.

\subsection{Mamba-Based Temporal Projector}

Traditional Video-LLMs often process video frames independently, requiring the LLM to infer temporal relationships from sequences of static images. This approach is computationally inefficient, particularly when processing long videos. Additionally, this method fails to leverage the inherent temporal redundancy in video data, resulting in redundant processing of similar information across consecutive frames. To address these limitations, we introduce the \textbf{Mamba-based temporal projector}, which efficiently integrates temporal information across video frames while enabling effective temporal token compression.

Let $\mathbf{X}_t \in \mathbb{R}^{\hat{N} \times D}$ denote the image tokens for frame $t$ by a Vision Transformer (ViT) encoder, where $\hat{N}$ is the number of tokens per frame and $D$ is the token dimension. We first apply a linear layer to downsample the tokens of each frame to $\frac{\hat{N}}{r}$ tokens:
\begin{equation}
\tilde{\mathbf{X}}_t = \text{Linear}\left( \mathbf{X}_t \right), \quad \text{for } t = 1, \dots, T.
\end{equation}

Where $r$ is the downsample ratio. For simplicity, we define $N = \frac{\hat{N}}{r}$ and use $N$ throughout the remainder of this paper. The downsampled tokens from all frames are stacked to form the input tensor for the temporal module:
\begin{equation}
\tilde{\mathbf{X}} = \left[ \tilde{\mathbf{X}}_1; \tilde{\mathbf{X}}_2; \dots; \tilde{\mathbf{X}}_T \right] \in \mathbb{R}^{T \times N \times D}.
\end{equation}

The core of the temporal projector consists of $L$ Mamba layers that iteratively integrate temporal dynamics into the tokens. In each layer $l = 1, \dots, L$, we fuse temporal information into the visual tokens by:
\begin{equation}
\mathbf{X}^{(l)} = \mathbf{X}^{(l-1)} + \text{MambaMixer}\left( \text{Norm}\left( \mathbf{X}^{(l-1)} \right) \right),
\end{equation}
where $\mathbf{X}^{(0)} = \tilde{\mathbf{X}}$, and $\text{Norm}(\cdot)$ denotes layer normalization. Each MambaMixer employs a bidirectional scanning module that captures dependencies in both spatial and temporal dimensions. Specifically, we apply a sweeping scan order within each frame and across frames, i.e., left-to-right, top-to-bottom, and frame-to-frame (see \autoref{fig:overview}). After $L$ layers, we obtain tokens enriched with temporal information, denoted as $\mathbf{X}^{(L)} \in \mathbb{R}^{T \times N \times D}$. 

\begin{figure*}[t]
    \centering
    \includegraphics[width=\textwidth]{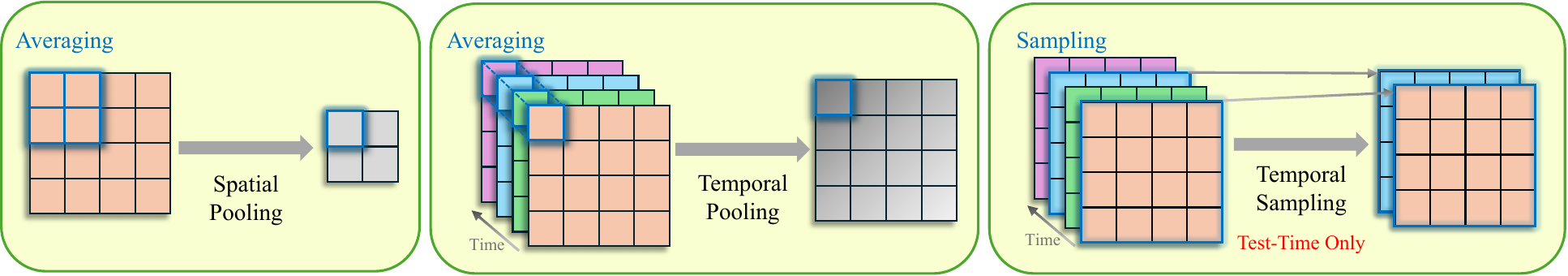}
    \caption{\textbf{Token Compression Strategies.} This figure illustrates our token compression techniques: spatial average pooling (left), temporal average pooling (middle), and training-free temporal token sampling (right). These methods can be applied individually or in combination, depending on task requirements and computational budget constraints. 
    }
    \label{fig:token_compression}
\end{figure*}

\subsection{Training-Time Token Compression}
Processing long videos poses two major challenges for LLMs. First, handling all frames is computationally expensive, often requiring specialized system optimizations like sequence parallelization and multiple GPUs for training and inference~\citep{longvila}. Second, LLMs are inherently limited by their training context lengths constraints. For example, LLaMA 3 has a context length of 8K tokens, which corresponds to only 32 frames for video inputs when using 256 tokens per frame. Without token compression, video processing quickly exceeds the effective input capacity of LLMs, leading to significantly reduced model performance. In this work, we aim to address both computational cost and context length limitations by enabling efficient long-video processing through token compression along both temporal and spatial dimensions. Our approach eliminates the need for custom system optimizations and allows inference on a single GPU. We illustrate the training-time token compression in \autoref{fig:token_compression} (left and middle).

\myparagraph{Temporal Pooling}
Since consecutive frames often contain similar information, analyzing every frame may lead to redundant processing and potential overfitting. Additionally, having too many tokens can make it difficult for the LLM to identify important temporal patterns. Thus, we propose applying temporal average pooling to compress visual information efficiently~\citep{lin2019tsm,wang2018tsn}. This method combines data from consecutive frames by averaging their enriched visual tokens. Specifically, for the tokens $\mathbf{X}^{(L)} \in \mathbb{R}^{T \times N \times D}$ from the temporal projector, we average every $k$ consecutive frames:

\begin{equation}
\mathbf{X}_{\text{time-avg}} = \frac{1}{k} \sum_{i=0}^{k-1} \mathbf{X}^{(L)}_{t+i}, \ \text{for } t = 0, k, 2k, \dots, T - k.
\end{equation}

As a result, we obtain compressed tokens:

\begin{equation}
\mathbf{X}_{\text{time-avg}} \in \mathbb{R}^{\frac{T}{k} \times N \times D},
\end{equation}

Despite its simplicity, temporal averaging effectively decreases the number of tokens the LLM processes, 
with minimal loss of critical information. 
This motivates us to adopt such simple yet effective technique for efficient training in long video understanding.

\myparagraph{Spatial Pooling}

In addition to temporal pooling, we also explore average pooling in the spatial domain. Formally, given the input $\mathbf{X}^{(L)} \in \mathbb{R}^{T \times N \times D}$ from the visual encoder and the spatial compression ratio $p$, we apply average pooling with a kernel size and stride of $p$ on each frame, resulting in $\mathbf{X}_{\text{space-avg}} \in \mathbb{R}^{T \times \frac{N}{p} \times D}$. 

\subsection{Training-Free Temporal Token Sampling}
\label{sec:temporal_token_subsampling}

After processing through the temporal projector, each visual token is enriched with spatiotemporal information, capturing features not only from its corresponding frame but also from other frames across the entire video. This encoding of global information allows us to subsample the visual tokens along the temporal dimension at test time, further reducing the number of tokens fed into the LLM without significant loss of information or performance. This temporal token subsampling strategy can be done with or without our pooling mechanisms. Note that comparing to methods that subsample raw frames and risk discarding critical temporal cues, our approach leverages the spatiotemporal richness of tokens post-temporal encoding. We show an illustration of the temporal token sampling in \autoref{fig:token_compression} (right).

Formally, let $\mathbf{X}^{'} \in \mathbb{R}^{T' \times N' \times D}$ denote the token input to the subsampling layer which can be tokens from visual encoder $\mathbf{X}^{(L)}$, or tokens output from compression modules $\mathbf{X}_{\text{time-avg}}$ or $\mathbf{X}_{\text{space-avg}}$. Here, $T'$ and $N'$ are the temporal and spatial dimensions after any temporal or spatial pooling.
We apply a uniform subsampling with rate $s$ along the temporal dimension.

\begin{align}
    \mathbf{X}_{\text{time-skip}} 
    = \{ \mathbf{X}^{'}_t \mid t = 0, k, & 2k, \dots, T-1 \} \notag \\
    &\in \mathbb{R}^{\frac{T'}{s} \times N' \times D} 
\end{align}

Our empirical results demonstrate that this subsampling method not only maintains performance on various video understanding benchmarks but can also improve it by reducing noise from redundant frames. 

\section{Experiments} \label{sec:experiments}

In this section, we extensively evaluate the proposed method on various video understanding benchmarks and provide empirical analysis demonstrating how the temporal projector enables efficient token reduction while delivering strong video reasoning abilities.

\subsection{Experiment Details} \label{sec:training_details}
We use the pre-trained vision encoder in PaliGemma~\citep{paligemma} and the LLM model from Qwen2-VL~\citep{wang2024qwen2vl}, and fine-tune them to adapt to our video datasets. The temporal projector is initialized with random weights. Each image is always resized to a $448 \times 448$ resolution. Note that we also carried out experiments on more model configurations in Section~\ref{sec:analysis} to showcase the universality of our design.

\begin{table*}
\setlength{\tabcolsep}{2.5pt}
\footnotesize\centering
\begin{adjustbox}{width=0.95\textwidth}
\begin{tabular}{lccccccccc}
\toprule
\multirow{2}{*}{\textbf{Models}} & \textbf{Size} & \textbf{Comp.} & \textbf{\#Frames }  & \textbf{\# Frames} & \textbf{MVBench} & \textbf{MLVU} & \textbf{LongVideoBench} & \textbf{VideoMME} \\
\cmidrule(lr){6-9}
  &   & \textbf{Ratio (\%)} & \textbf{(train)} & \textbf{(test)} & \textbf{test}  & \textbf{dev} & \textbf{val} & \textbf{(w/o sub.)} \\
\midrule
\textbf{Duration} & & &  &  & 16 sec & 3$\sim$120 min & 8 sec$\sim$60 min & 1$\sim$60 min \\
\midrule
\textbf{Proprietary Models} & & & &  &  &  &  &  \\
GPT4-V & - & - &  & 1fps  & 43.7 & - & - & 60.7 \\
GPT4-O & - & - & & 1fps  & 64.6 & 66.2 & - & 77.2 \\
\midrule
\textbf{Open-Source Video-LLMs} & & & &   &  &  &  &  \\
Video-LLaVA~\citep{lin2023videollava}   & 7B & - & 8 & 8  & 41.0 & 47.3 & - & 40.4 \\
LLaMA-VID~\citep{li2023llamavid}   & 7B & - & 1fps & 1fps  & 41.9 & 33.2 & - & - \\
Chat-UniVi~\citep{jin2023chatunivi}   & 7B & - & 64 & 64  & - & - & - & 45.9 \\
ShareGPT4Video~\citep{chen2024sharegpt4video}   & 8B & - & 16 & 16  & 51.2 & 46.4 & - & 43.6 \\
LLaVA-NeXT-Video~\citep{zhang2024llavanextvideo}   & 7B & - & 16 & 32  & 33.7 & - & - & 46.5 \\
VideoLLaMA2~\citep{damonlpsg2024videollama2}   & 7B & - & 16 & 32 & 54.6 & 48.5 & - & 46.6 \\
VideoChat2~\citep{li2023videochat}   & 7B & - & 8 & 16  & 60.4 & 47.9 & - & 54.6 \\
Long-LLaVA~\citep{wang2024longllava}  & A13B (53B) & - & - & 128 & 54.6 & - & - & 51.6 \\
LongVA~\citep{zhang2024longva}  & 7B & - & - & 128  & - & 56.3 & - & 54.3 \\
LongVILA~\citep{longvila}  & 7B & - & 2048 & 256  & - & - & - & 57.5 \\
mPLUG-Owl3~\citep{ye2024mplugowl3} & 8B & - & 8 & 128  & 54.5 & - & 52.1 & 53.5 \\
LLaVA-OneVision~\citep{li2024llavaonevision}  & 7B & - & 32 & 32  & 56.7 & 64.7 & {56.5} & 58.2 \\
Kangaroo~\citep{liu2024kangaroo}  & 8B & - & 160 & 64  & 61.0 & 61.1 & 54.8 & 56.0 \\
VideoXL~\citep{shu2024videoxl}  & 7B & - & 128 & 2048  & 55.3 & 64.9 & 49.5 & 55.5 \\
Oryx-1.5~\citep{liu2024oryx}  & 7B & - & 256 & 64  & 67.6 & 67.5 & {56.3} & 58.8 \\
LongVU~\citep{shen2024longvu}  & 7B & - & - & 1fps  & 66.9 & 65.4 & - & 60.6 \\
Qwen2-VL~\cite{wang2024qwen2vl} & 7B & - & 2fps & 2fps & 67.0 & - & 55.6 & 63.3 \\
\midrule
\rt \textbf{Ours} & 7B & \textbf{25/12.5}* & 128 & 256  & \textbf{71.3} & \textbf{72.9}* & \textbf{60.5}* & \textbf{63.4} \\
\bottomrule
\multicolumn{9}{l}{\small * test-time temporal sampling is applied, 2$\times$ additional compression.} \\
\end{tabular}
\end{adjustbox}
\caption{\textbf{Comparison with Existing Video-LLMs.} We compare our best-performing configuration (\method + T. Pooling) against existing Video-LLMs. Our approach outperforms all existing models while using significantly fewer tokens through $4 \times$ temporal pooling token compression. Additionally, values with asterisk (*) indicate that test-time token sampling is applied, which introduces additional $2 \times$ compression on visual inputs. Latency are based on 256 frames. \# Frame (test) denote the maximum frames used in all benchmarks.}
\label{tab:main_results_compare_external}
\end{table*}
\myparagraph{Training}
In the first stage, known as the Alignment Stage, we freeze both the image encoder and the LLM, training only the temporal projector using a small image-text dataset \citep{zhao2023svit}, containing 95K image-text pairs. 
Note that the Mamba layers perform not only temporal scan but also spatial scan within images, so video inputs are not strictly required to train it. For alignment stage, we find it sufficient to use only image-text pairs to pretrain the temporal projector. In the second stage, the supervised fine-tuning stage (SFT), we fine-tune all three components using a large and diverse dataset that includes text-only, image-text, and video-text data. There are around 12.5M samples in our SFT data mixture.
due to space constraints.
At this stage, we use 32 frames for each video input. For models with training-time token compression
, we use a compression ratio of 4$\times$ --- temporal pooling models compress 32 frames to 8 frames while spatial pooling models compress 256 tokens per image into 64 tokens. Moreover, for models with training-time token compression 
, we further employ a long video fine-tuning stage using 128-frames long-video inputs from the 
LLaVA-Video dataset~\citep{zhang2024llavavideo}. We provide further details about the full SFT dataset and long video fine-tuning dataset in the appendix Section~\ref{sec:appx_dataset}.

\begin{table*}[t]
\setlength{\tabcolsep}{1.3pt}
\footnotesize\centering
\begin{adjustbox}{width=0.95\textwidth}
\begin{tabular}{lccccccccc}
\toprule
\multirow{2}{*}{\textbf{VILA-Based Models}} & \textbf{Size} & \textbf{Comp.} & {\textbf{Latency}} & \textbf{\#Frames }  & \textbf{\# Frames} & \textbf{MVBench} & \textbf{MLVU} & \textbf{LongVideoBench} & \textbf{VideoMME} \\
\cmidrule(lr){7-10}
  &   & \textbf{Ratio (\%)} & \textbf{(s)} & \textbf{(train)} & \textbf{(test)} & \textbf{test}  & \textbf{dev} & \textbf{val} & \textbf{(w/o sub.)} \\
\midrule
\textbf{Duration} & & &  & &  & 16 sec & 3$\sim$120 min & 8 sec$\sim$60 min & 1$\sim$60 min \\
\midrule
\textbf{Token Budget: 8K} \\
VILA Baseline & 7B & 100 & 4.31 & 32 & 256 & 69.5 & 70.2 & 55.9 & 60.1 \\
\rc \method  & 7B & 100 & 4.47 & 32 &  256  & {70.3} & {71.1} & 54.5 & 62.4 \\
\rc \method+ T. Pooling & 7B & 25 & 1.82 & 128 & 256  & \textbf{71.3} & {72.5} & 59.5 & \textbf{63.4} \\
\rt \method  + T. Sampling *  & 7B & 50 & 2.50 & 32 & 256 & {70.1} & 70.8 & 54.8 & {63.1} \\
\rt \method+ T. Pooling + T. Sampling * & 7B & 12.5 & 1.51 & 128 & 256 & 70.6 & \textbf{72.9} & \textbf{60.5} & {62.4} \\
\bottomrule
\multicolumn{10}{l}{\small * 2$\times$ additional compression at test time.} \\
\end{tabular}
\end{adjustbox}
\caption{\textbf{Comparison between VILA-Based Models on the Same Token Budget.} 
We compare between \method variants and the baseline VILA model, all trained with identical data and training pipelines. All models use the same 8K token budget, which is the maximum number of visual tokens provided to the LLM during training. Temporal pooling (T. Pooling) applies $4 \times$ compression and test-time token sampling (T. Sampling) applies additional $2 \times$ compression.}
\label{tab:main_results_compare_internel}
\end{table*}

\myparagraph{Evaluation}

We evaluate our \method across multiple configurations on recent long-video understanding benchmarks specifically designed to rigorously assess video-language model capabilities. These benchmarks include 
MVBench~\citep{li2023mvbench}, MLVU~\citep{zhou2024mlvu}, LongVideoBench~\citep{wu2024longvideobench}, and VideoMME~\citep{fu2024videomme}. We compare our approach against a wide range of representative video-language models, including recently proposed models tailored for long-video understanding~\citep{longvila, shen2024longvu, zhang2024longva, shu2024videoxl, wang2024longllava, liu2024kangaroo, liu2024oryx}. See \autoref{tab:main_results_compare_external} for details. 

\myparagraph{Models} We implement \method based on the VILA codebase~\cite{lin2024vila}, a typical VLM pipeline consisting of a vision encoder, LLMs, and vision-language projector, and introduce our novel Mamba module and compression mechanisms into the architecture. We will refer all models trained with the VILA codebase as VILA-based models in our experiments. To thoroughly analyze our design, we evaluate \method variants using all combinations of the three compression methods: temporal average pooling, spatial average pooling, and temporal token sampling. The full results are presented in \autoref{tab:comp_abl}. For comparison with existing Video-LLMs, we highlight the best-performing variants in \autoref{tab:main_results_compare_external}. And include the detailed analysis of all variants of \method in \autoref{tab:main_results_compare_internel}. To ensure fairness, we also include a baseline VILA model trained on the same dataset and training scheme but without the Mamba module. All of our models in \autoref{tab:main_results_compare_internel} are trained under the same 8K token budget (corresponding to 32 frames of tokens 
), which represents the number of visual tokens fed to the LLM—a key factor affecting inference latency and memory consumption, particularly as the number of frames increases (see Section~\ref{sec:analysis}). Specifically, we report results for the standard \method trained on 32-frame inputs and \method with Temporal Pooling, which processes 128-frame inputs while reducing the token count to match the 32-frame variant. Additionally, we evaluate configurations where temporal sampling is applied at test time (+T. Sampling), which not only further enhances model efficiency but also improves performance on certain benchmarks. 

\subsection{Results on Video Understanding Benchmarks} \label{sec:benchmark_results}
\myparagraph{\method vs Existing Methods}
We start with comparing our configuration \method + T. Pooling (+ T. Sampling) with existing Video-LLMs. As shown in \autoref{tab:main_results_compare_external} and detailed in \autoref{tab:main_results_compare_internel}, \method + T. Pooling achieves new state-of-the-art performance across all long-video understanding benchmarks. Specifically, it achieves 71.3\% accuracy on MVBench, 72.5\% on MLVU, 59.5\% on LongVideoBench, and 63.4\% on VideoMME, outperforming all open-source Video-LLMs, including recent models specifically designed for long-context inputs such as LongVU and LongVILA. Additionally, our method significantly narrows the performance gap with proprietary models, outperforming GPT4-V and GPT4-O on MVBench and MLVU, as well as GPT4-V on VideoMME. Notably, \method + T. Pooling achieves computational efficiency by compressing visual tokens to 25\% of their original number before processing them through the LLM. We can further enhance efficiency by applying test-time temporal sampling in \method + T. Pooling + T. Sampling while still achieving competitive results, reducing the token count to just 12.5\% while maintaining competitive performance. In fact, this additional compression even improves results of certain benchmarks, yielding the best overall performance on MLVU and LongVideoBench.

\myparagraph{\method vs Baseline VILA} 
Next, we provide controlled comparisons within VILA-based models to reveal advantages of the proposed Mamba-based temporal module in \autoref{tab:main_results_compare_internel}. We first compare the baseline VILA model with our \method.
By incorporating the Mamba module, \method achieves performance gains on three out of four benchmarks, including a notable 2.3\% improvement on VideoMME. Moreover, augmenting \method with test-time temporal token sampling (\method + T. Sampling) further enhances efficiency, reducing inference time by 43\% while, surprisingly, maintaining or slightly boosting performance (an additional 0.8\% gain on VideoMME). This advantageous behavior emerges because the Mamba module’s ability to effectively propagate temporal information across video frames, enabling redundant tokens to be discarded without compromising the model’s overall understanding.

The temporal pooling variant (\method + T. Pooling) extends these benefits to long-context training, by applying temporal average pooling after the Mamba layer, which allows the model to process 128-frame inputs while compressing the token count to match that of the 32-frame setting. This approach not only improves performance, achieving 63.4\% (+3.3\%) on VideoMME, 59.5\% (+3.6\%) on LongVideoBench, 72.5\% (+2.3\%) on MLVU, and 71.3\% (+1.8\%) on MVBench, but also significantly reduces inference latency by 58.4\%. 
By combining this model with test-time temporal token sampling (\method + T. Pooling + T. Sampling), we further reduce the inference time by 65.5\% and use only 12.5\% of the visual tokens compared to VILA without sacrificing performance. 
We observe that this test-time temporal token sampling particularly benefits MLVU and LongVideoBench compared to MVBench and VideoMME. This different impact across benchmarks likely stems from the nature of the underlying tasks. MLVU and LongVideoBench require global understanding across long videos. We believe that test-time compression with the Mamba module better summarizes the essential contextual information. On the other hand, MVBench and VideoMME require visual details from specific frames. Our pooling-only method with the Mamba module maintains more detailed frame information throughout the sequence. Section \ref{sec:analysis} includes a detailed discussion about retaining visual information with our token compression.

\myparagraph{Spatial Pooling vs Temporal Pooling on Long Video Inputs} \autoref{tab:spatial_vs_temporal_compression} provide comparison between spatial average pooling and temporal average pooling on 32 frames training and 128-frames extension. All models use \method as the base model. We see that spatial pooling is effective when 32 frames are during training, outperforming temporal pooling and on LongVideoBench while achieving on par results on VideoMME. However, when applying 128-frames input, even though both methods use the same token budget, temporal pooling results in sigificantly better performance. In fact, spatial pooling can not benefit from longer video inputs, results in degrated performance on both LongVideoBench and VideoMME, while temporal pooling successfully achieves stronger performance on both benchmarks from the extended video length. 

\begin{table}
    \centering
    \footnotesize
    \begin{adjustbox}{width=0.9\linewidth}
    \begin{tabular}{lcccc}
        \toprule
        \textbf{Models}& \multicolumn{2}{c}{\textbf{LongVideoBench}} & \multicolumn{2}{c}{\textbf{VideoMME}} \\
         \cmidrule(lr){2-3} \cmidrule(lr){4-5}
          & \textbf{32F} & \textbf{128F} & \textbf{32F} & \textbf{128F} \\
        \midrule
        S. Pooling & \textbf{56.0} & 55.9(\textcolor{red}{-0.1}) & 61.1 & 58.3 (\textcolor{red}{-2.8})\\
        T. Pooling & 54.2 & \textbf{59.5} (\textcolor{blue}{+5.3}) & \textbf{61.2}  & \textbf{63.4} (\textcolor{blue}{+2.2}) \\
        \bottomrule
    \end{tabular}
    \end{adjustbox}
    \caption{\textbf{Spatial Pool vs Temporal Pooling on Frame Extension.} Models were initially trained on 32 frames, then fine-tuned on 128 frames using a small long video dataset (see Section \ref{sec:appx_dataset}). F indicates frame count during training. Both methods compress visual tokens to 25\% of the original count. Spatial pooling performance degrades with increased frame length, while temporal pooling shows consistent improvements.}
    \label{tab:spatial_vs_temporal_compression}
\end{table}

\section{Analysis} \label{sec:analysis}

\begin{table}
    \centering
    \footnotesize
    \setlength{\tabcolsep}{2pt}
    \begin{adjustbox}{width=0.9\linewidth}
    \begin{tabular}{lccc}
        \toprule
        \textbf{Models} & \textbf{8F} & \textbf{32F (T. Pooling)} & \textbf{128F (T. Pooling)} \\
        VILA (w/o Mamba) & \textbf{62.0} & 63.5 (\textcolor{blue}{+1.5})  & 64.3 (\textcolor{blue}{+0.8}) \\
        \method (w/ Mamba) & 61.6 & \textbf{64.2} (\textcolor{blue}{+2.6}) & \textbf{66.7} (\textcolor{blue}{+2.5}) \\
        \bottomrule
    \end{tabular}
    \end{adjustbox}
    \caption{\textbf{The Critical Role of the Mamba Module.} Results show average performance across all video benchmarks. Numbers in parentheses show improvements over shorter version of the same models. For 32F and 128F results, models were initially trained on the full dataset with 32 frame inputs, then fine-tuned on 128 frame inputs using a smaller long-video dataset (see Section \ref{sec:appx_dataset}). \method demonstrates substantially better utilization of longer temporal contexts, with the performance gap between VILA baseline and \method widening as input length increases (+0.7\% at 32F and +2.4\% at 128F compared to baseline). Full details of the comparisons are provided in \autoref{tab:mamba_compression_full}.}
    \label{tab:mamba_compression}
\end{table}

\myparagraph{Mamba Module Improves Simple Token Compression} 
As illustrated in \autoref{fig:overview}, the Mamba-based temporal projector performs a spatiotemporal scan over input tokens. This enables refinement of visual tokens before compression operations, thereby preserving key temporal and spatial cues—such as the positional information of frames being pooled—that would otherwise be lost in naive compression strategies. Consequently, the Mamba temporal module allows simple token compression methods to work more effectively for long videos compared to the baseline, as evidenced in \autoref{tab:mamba_compression} and further detailed in \autoref{tab:mamba_compression_full}.

Our results demonstrate that \method consistently improves with increased video input length during training, achieving substantial gains of +2.6\% over all benchmarks when extending from 8 frames to 32 frames and an additional +2.5\% when extending from 32 frames to 128 frames. In contrast, the baseline VILA model shows smaller improvement when extending from 8 frames to 32 frames, and demonstrates only +0.8\% performance gains when fine-tuned on 128 frames. In fact, as shown in \autoref{tab:mamba_compression_full}, the baseline model exhibits 
performance degradation on MVBench and MLVU benchmarks when extending video length from 32 frames to 128 frames. These results fully demonstrate the importance of the Mamba module in enabling both efficient and effective token compression, particularly for long-video input
\begin{table*}
\setlength{\tabcolsep}{1.5pt}
\small\centering
\begin{adjustbox}{width=0.95\textwidth}
\begin{tabular}{l|c|ccc|ccc|cccccc}
\toprule
\multirow{2}{*}{\textbf{Model}} & \textbf{Mamba} & \textbf{T.} & \textbf{T.} & \textbf{S.}  & \textbf{Token} &  \textbf{Compression}  & \multirow{2}{*}{\textbf{Latency (s)}} & \multirow{2}{*}{\textbf{MVBench}} & \multirow{2}{*}{\textbf{MLVU}} & \multirow{2}{*}{\textbf{LongVideoBench}} & \textbf{VideoMME } \\
& \textbf{Module} & \textbf{Sampling} & \textbf{Pooling} & \textbf{Pooling} & \textbf{budget} & \textbf{Ratio (\%)}  & & & & & \textbf{w/o sub.}\\
\midrule
VILA & \textcolor{gray}{\ding{55}} & \textcolor{gray}{\ding{55}} & \textcolor{gray}{\ding{55}}  & \textcolor{gray}{\ding{55}} & 8K & 100 & 4.31  & 69.5 & 70.2  & 55.9 & 60.1 \\ 
\midrule
\multirow{8}{*}{\method}   & \ding{51} & \textcolor{gray}{\ding{55}} & \textcolor{gray}{\ding{55}} & \textcolor{gray}{\ding{55}} & 8K & 100 & 4.47 & \textbf{70.3} & \textbf{71.1}  & 54.5 & 62.4 \\ 
    & \ding{51} & \ding{51}  & \textcolor{gray}{\ding{55}} & \textcolor{gray}{\ding{55}} & 8K & 50  & 2.50 & {70.1} & 70.8  & 54.8  & \textbf{63.1} \\ 
    \cmidrule(lr){2-12}
    & \ding{51} & \textcolor{gray}{\ding{55}} & \ding{51} & \textcolor{gray}{\ding{55}} & 2K & 25  & 1.82 & \textbf{70.4}  & \textbf{71.0}  & 54.2 & {61.2}  \\
    & \ding{51} & \textcolor{gray}{\ding{55}} & \textcolor{gray}{\ding{55}} & \ding{51} & 2K & 25  & 1.82 & 68.9  & 69.2 & \textbf{56.0} & 61.1 \\
    \cmidrule(lr){2-12}
    & \ding{51} & \ding{51} & \ding{51} & \textcolor{gray}{\ding{55}} & 2K & 12.5 & 1.51 & \textbf{70.1} & \textbf{71.0} & 54.0 & 60.8  \\ 
    & \ding{51} & \ding{51}& \textcolor{gray}{\ding{55}} & \ding{51} & 2K & 12.5 & 1.51 & 68.9 & 69.5 & \textbf{56.3} & 60.9 \\
    \cmidrule(lr){2-12}
    & \ding{51} & \textcolor{gray}{\ding{55}} & \ding{51} & \ding{51} & 0.5K & 6.25 & 1.36 & 68.5   & 68.2  & 53.7 & 60.2 \\
    & \ding{51} & \ding{51} & \ding{51} & \ding{51} & 0.5K & 3.13 & 1.29 &  67.2 & 68.6 & 55.3 & 60.1 \\
\bottomrule
\end{tabular}
\end{adjustbox}
\caption{\textbf{Ablation of token compression methods.} All models are trained on 32 frames. The pooling methods are added during training, and the temporal sampling is executed at test-time.}
\label{tab:comp_abl}
\end{table*}

\myparagraph{Ablation of Mamba and token compression modules}
\autoref{tab:comp_abl} presents an ablation study comparing various token compression strategies for models trained on a fixed 32-frame input. Without any compression, our standard \method delivers improved performance across benchmarks comparing to the baseline VILA while utilizing a similar inference latency and number of visual tokens. When test-time temporal sampling is applied, the model maintains its performance while reducing the visual token count to 50\% and lowering inference latency to 58.7\% of the uncompressed \method. Employing temporal average pooling during both training and testing further compresses the token count to 25\% and cuts latency down to 42.7\%, which is particularly effective for MVBench and MLVU but slightly degrades LongVideoBench and VideoMME. Similarly, spatial average pooling provides the same efficiency improvements and proves particularly effective on LongVideoBench, but with compromises on other benchmarks. 

We further apply temporal token sampling with either temporal or spatial pooling to improve the test-time efficiency. For instance, \method + T. Pooling + T. Sampling reduces the visual tokens to only 12.5\% and latency to 35.4\% of the original \method, while maintaining comparable accuracy to \method + T. Pooling. The \method + S. Pooling + T. Sampling even achieves the best LongVideoBench performance with this strong token reduction. Finally, we combine the temporal and spatial average pooling with/without test-time temporal token sampling, leading to even more aggressive compression. Interestingly, our \method + T. Pooling  + S. Pooling + T. Sampling variant, which has the strongest compression ratio, utilizing only 3.13\% visual token and 29.5\% of inference latency, already achieves a competitive results. Comparing to the baseline VILA, it achieves 100\% performance on VideoMME, 98.9\% on LongVideoBench, 97.7\% on MLVU, and 96.7\% on MVBench, making it quite attractive for efficiency-oriented scenarios.

We note that, although the uncompressed \method and its test-time sampling variant achieve the highest overall performance when using the same 32 frames input, their computational demands limit scalability to train on longer sequence. In contrast, as demonstrated in \autoref{tab:main_results_compare_internel}, 
the \method + T. Pooling allows extending the temporal context to 128 frames with temporal pooling, which enables improved performance without requiring additional LLM computation.

\begin{figure*}[!htbp]
\centering
\begin{subfigure}[b]{0.34\textwidth}
    \centering
    \includegraphics[width=\textwidth]{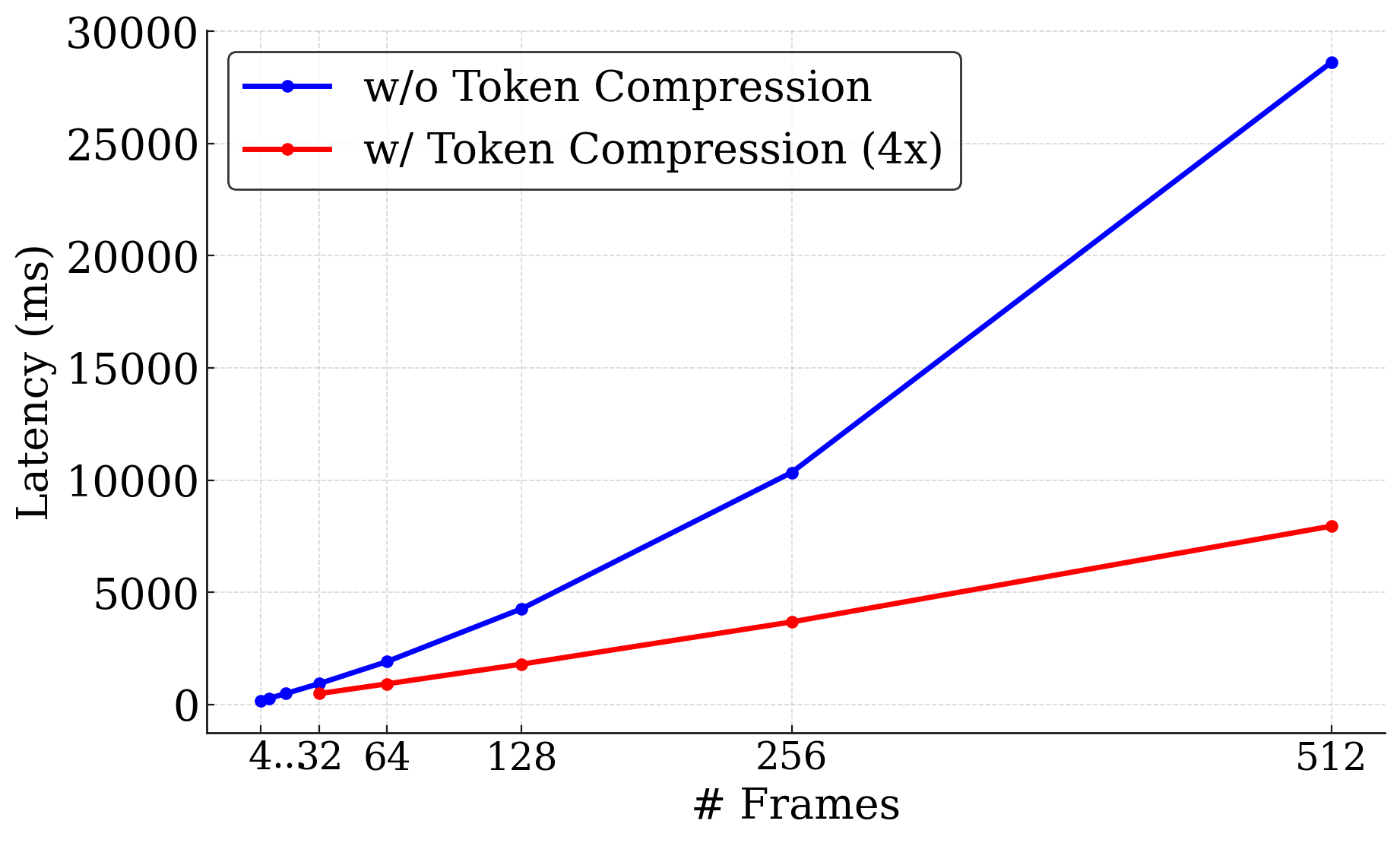}
    \label{fig:speedup}
\end{subfigure}
\hfill
\begin{subfigure}[b]{0.27\textwidth}
    \centering
    \includegraphics[width=\textwidth]{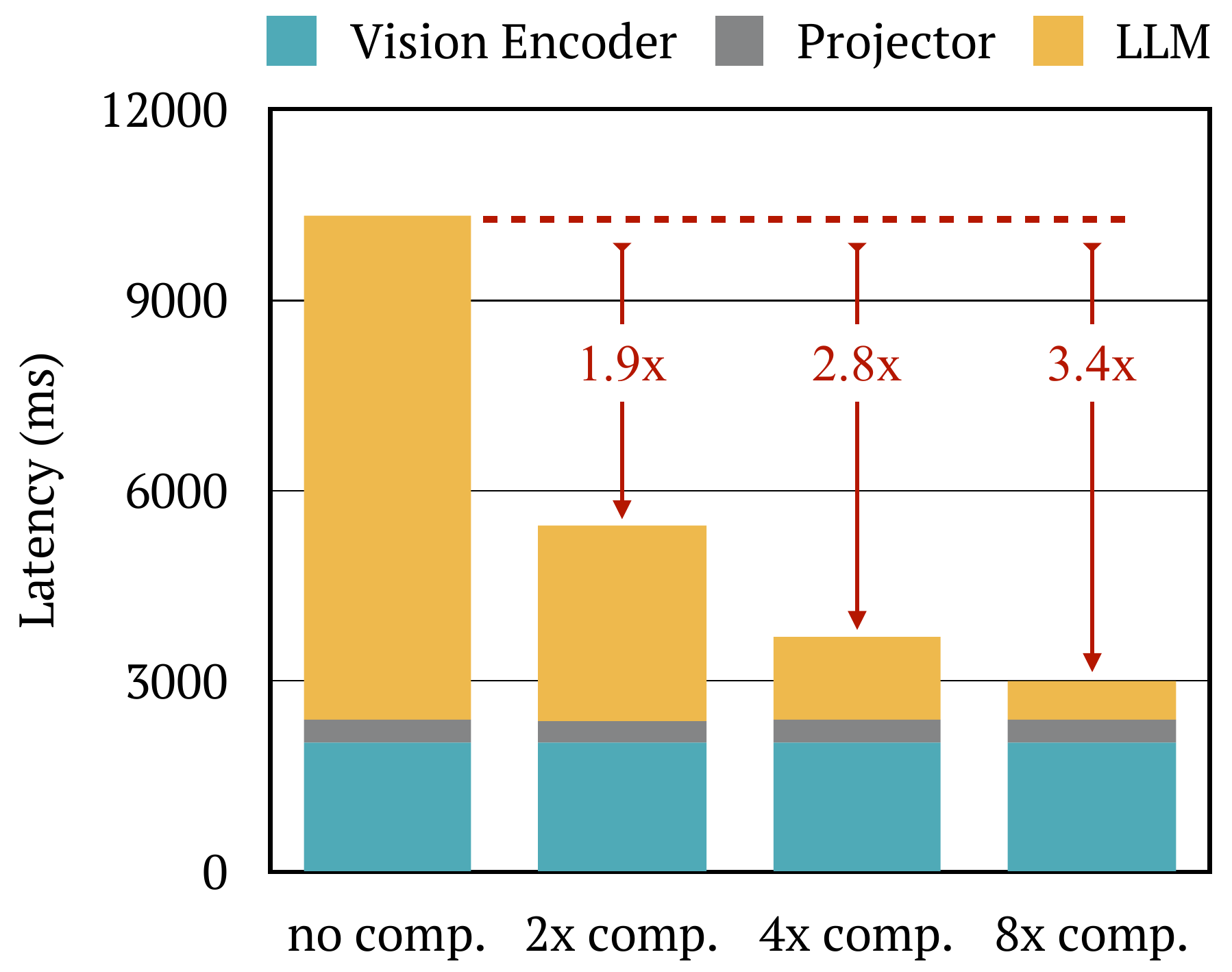}
    \label{fig:latency}
\end{subfigure}
\hfill
\begin{subfigure}[b]{0.34\textwidth}
    \centering
    \includegraphics[width=\textwidth]{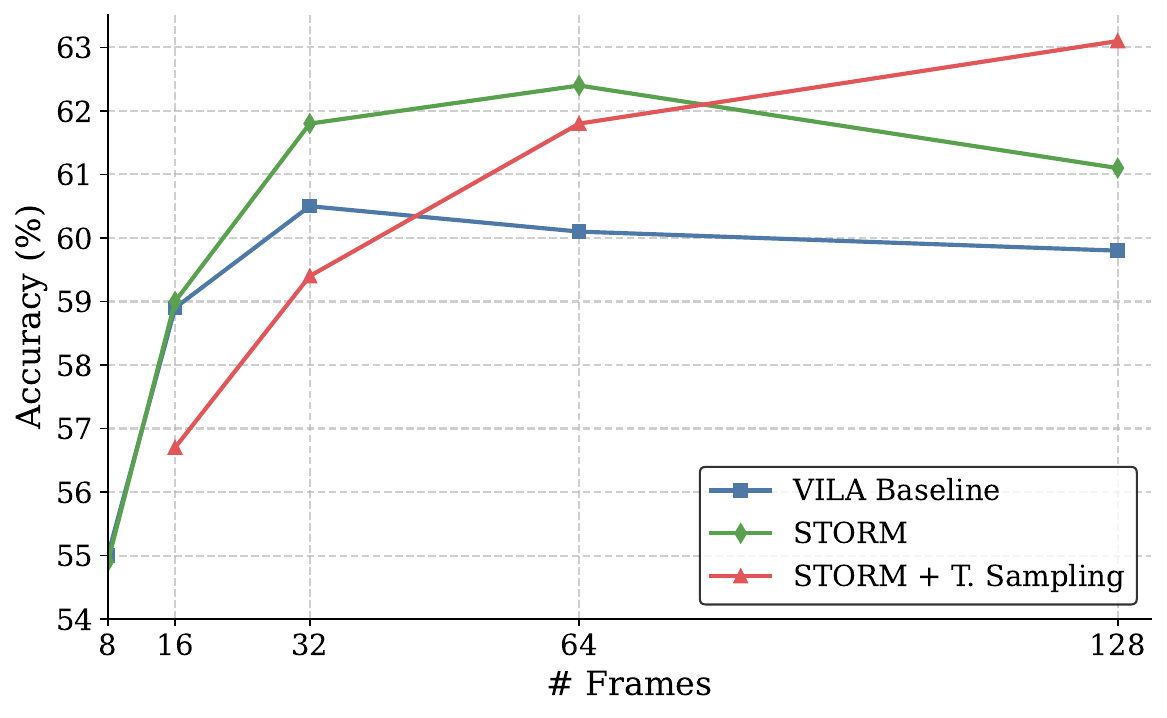}
    \label{fig:framewise}
\end{subfigure}
\caption{\textbf{Model Efficiency and Effectiveness on Long Video Inputs.} \textbf{(left)} Profiling results of token compression as the number of frames increases during inference. \textbf{(middle)} Profiling results for 256 input frames with different compression ratios on a single A100. \textbf{(right)} The accuracy of Video-MME (without subtitles) across different numbers of frames during inference. While \method with test-time temporal sampling showed consistent performance improvements, both VILA and \method without token compression demonstrated decreased performance beyond 64 frames.}
\label{fig:efficiency}
\end{figure*}

\begin{figure*}[!htbp]
    \centering
    \includegraphics[width=\textwidth]{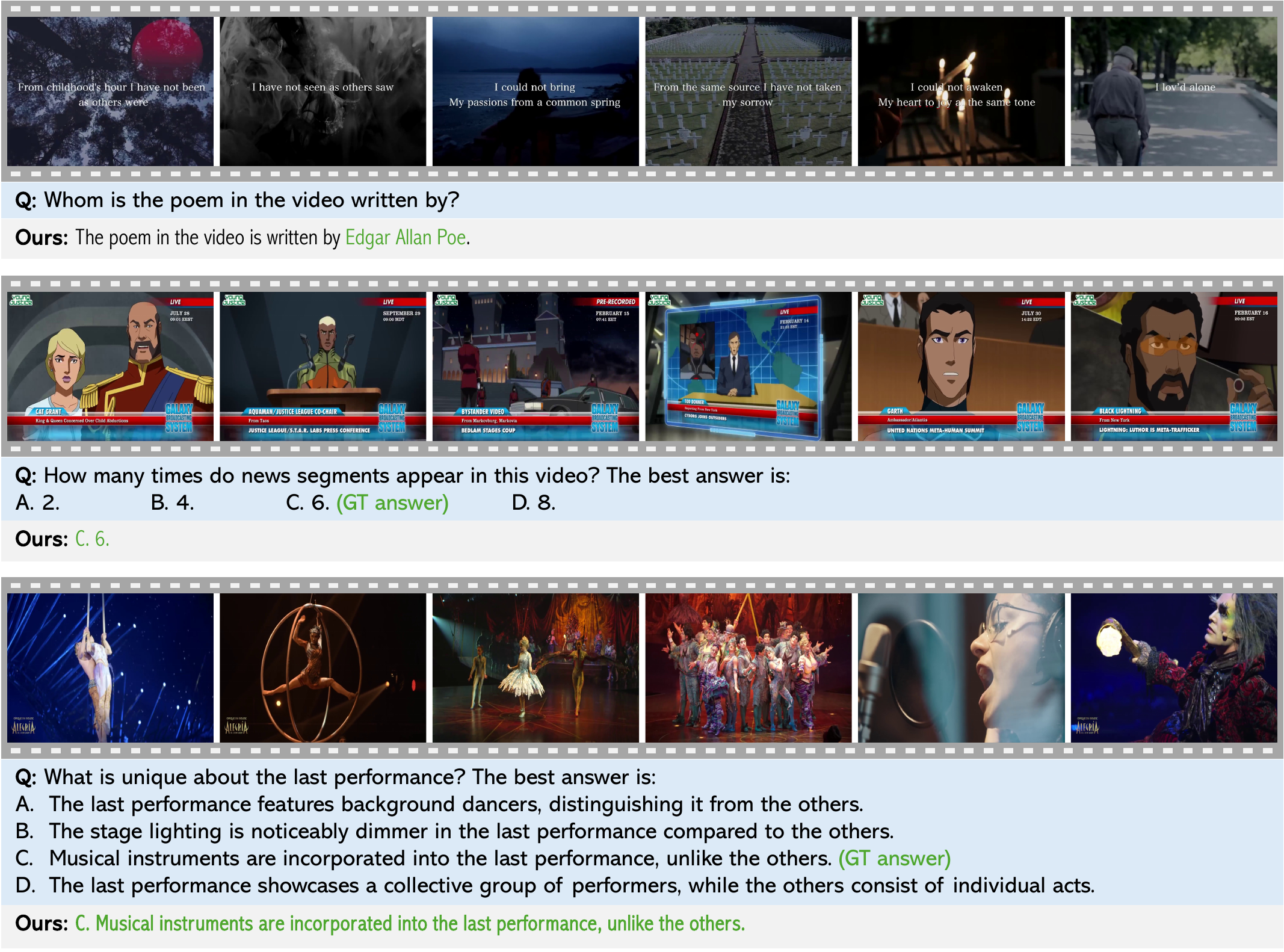}
    \caption{\textbf{Qualitative Examples of \method + T. Pooling.} Our model effectively processes complex video content across various tasks requiring fine-grained temporal and visual understanding while reducing computational overhead through efficient token compression. The example videos can be found in \href{https://research.nvidia.com/labs/lpr/storm}{our website}.}
    \label{fig:retrain_visual_info}
\end{figure*}

\myparagraph{Token Compression Improves Model Efficiency}
We provide analysis on model efficiency in \autoref{fig:efficiency}. In \autoref{fig:efficiency} left, we compare the inference latency of \method with and without token compression and across varying numbers of input frames. Due to the quadratic computation of the LLM, the latency gap between the two configurations widens significantly as the sequence length increases. \autoref{fig:efficiency} middle provides a detailed breakdown of inference latency of different modules for processing 256 frames. Without token compression, approximately 80\% of the total latency is attributed to the LLM module. By applying a token compression ratio of 4, the latency associated with the LLM is reduced to roughly 35\% of its original value, demonstrating a substantial improvement in efficiency.

Importantly, the temporal projector enables token compression without compromising performance. In fact, for some benchmarks, it even improves model performance while reducing computation cost, as demonstrated in \autoref{tab:main_results_compare_internel} and detailed in \autoref{tab:main_results_full}. 
\autoref{fig:efficiency} right presents a detailed analysis using the temporal sampling method on the VideoMME benchmark. We compare three configurations: the VILA baseline, \method without compression, and \method with test-time temporal token sampling 
Using up to 128 input frames and a compression ratio of 2. The results indicate that: (1) The VILA baseline's performance improves up to 32 frames but declines beyond this point, (2) \method generalizes better to longer sequences, maintaining performance up to 64 frames before a slight drop, and (3) \method with temporal sampling continues to improve performance up to 128 frames while providing a 50\% token compression ratio, meaning \method + T. Sampling at 128 frames uses the same number of visual tokens as the other two models at 64 frames.

We speculate that phenomenon is because of the limitation of the pre-trained LLM models which, despite having large context lengths on paper, operate with a smaller effective context length in practice. This limitation hinders their performance on very long sequences. The combination of the temporal projector and temporal sampling ensures that the number of tokens fed into the LLM remains within its effective context length. Simultaneously, the enriched visual tokens retain comprehensive temporal information beyond the sampled frames.

\myparagraph{Mamba Module Has Minimal Training Overhead}
The Mamba projector introduces minimal training overhead: even when no compression is applied, it adds only $\sim$ 5$\%$ latency in full training (19.1 hours with VILA (without Mamba module) and 20.1 hours with \method (with Mamba module)). More importantly, since the Mamba module enables effective token compression, it in fact provides significant training / inference speedups and performance gains (e.g., +3.6$\%$ on MLVU with 4× fewer tokens), eventually reducing training the costs.

\myparagraph{Our Token Compression Retains Visual Information}

The qualitative results in \autoref{fig:retrain_visual_info} demonstrate that \method's token compression preserves critical visual information while significantly reducing computational overhead. Even with 4$\times$ compression ratio, our model accurately extracts task-relevant information across diverse video understanding tasks. For example, questions like 'Whom is the poem in the video written by?', 'How many times do news segments appear in this video?', and 'What is unique about the last performance?' require fine-grained visual information and focus on the specific content from the prompt.
Detailed category-level analysis on VideoMME (\autoref{fig:videomme_category}) also reveals that \method + token compression consistently outperforms baseline model across nearly all task categories. Even in OCR-heavy tasks that require detailed visual analysis, our compressed model maintains performance comparable to uncompressed versions while using only a fraction of the computational resources (only 25\% visual tokens).
These findings demonstrate that our Mamba-based temporal encoder enables effective token compression by encoding spatiotemporal relationships directly into visual tokens, allowing the model to maintain high-level understanding while drastically reducing token count. Additional qualitative results are provided in \autoref{fig:appx_qualitative_num_input_frames}-\ref{fig:appx_qualitative_categories_showcase_3}.

\myparagraph{Our Modules Generalize to Various Architectures}
Our core design, temporal module + token compression, is model-agnostic and can be integrated into various video-LLM architectures. \autoref{tab:architecture} shows the experiments on models utilizing different LLM models (Qwen2~\cite{wang2024qwen2vl} vs Llama3~\cite{llama3}), vision encoders (PaliGemma~\cite{paligemma} vs SigLip~\cite{zhai2023sigmoid}), model sizes (8B vs 8B vs 1.5B), and input resolutions ($448\times448$ vs $384\times384$). Our design shows consistent improvements in both performance and latency for all configurations, clearly showcasing the universality of the model.

\begin{table}
    \centering
    \footnotesize
    \setlength{\tabcolsep}{3pt}
    \begin{adjustbox}{width=0.95\linewidth}
    \begin{tabular}{lccccc}
        \toprule
            \multirow{2}{*}{\textbf{Architectures}} & \multirow{2}{*}{\textbf{Resolution}}  & \multicolumn{2}{c}{\textbf{Latency (ms)}} & 
            \multicolumn{2}{c}{\textbf{VideoMME}} \\
            \cmidrule(lr){3-4} \cmidrule(lr){5-6} 
            & & {VILA} & {Ours} & {VILA} & {Ours}  \\
        \midrule
        Qwen2 7B + PaliGemma & 448$\times$448 & 1980 & \textbf{926} & 59.7 & \textbf{61.2} \\
        Llama3 8B + SigLip & 384$\times$384 & 1560 & \textbf{724} & 54.6 & \textbf{56.8} \\
        Qwen2 1.5B + SigLip & 384$\times$384 & 724 & \textbf{486} & 49.6 & \textbf{52.6} \\
        \bottomrule
    \end{tabular}
    \end{adjustbox}
    \caption{\textbf{Performance Comparison across Model Architectures, Model Sizes, and Input Resolutions.} We provide results of benchmark VideoMME, without subtitles version. 
    }
    \label{tab:architecture}
\end{table}

\begin{table}
    \centering
    \footnotesize
    \begin{adjustbox}{width=0.95\linewidth}
    \begin{tabular}{lcc}
        \toprule
        \textbf{Models}  & \textbf{32F (T. Pooling)} & \textbf{128F (T. Pooling)} \\
        VILA & 58.9  & 61.7 \\
        Uni-dir \method  & \textbf{62.2}  & 62.5 \\
        Bi-dir \method  & 61.2 & \textbf{63.4} \\
        \bottomrule
    \end{tabular}
    \end{adjustbox}
    \caption{\textbf{Support for Streaming/Online Settings.} We evaluate a uni-directional variant of \method designed for streaming video inputs. Results show that the Uni-dir \method consistently outperforms the VILA baseline, highlighting the potential of our design to support streaming scenarios.}
    \label{tab:streaming_support}
\end{table}

\myparagraph{Support for Streaming/Online Settings.}
To evaluate the applicability of our method in streaming scenarios, we replaced the default bi-directional Mamba with a uni-directional variant, allowing the model to reuse prior states for constant-time computation as new frames arrive. The results are provided in \autoref{tab:streaming_support}. We find that both uni-directional and bi-directional variants significantly outperform the baseline without temporal modeling. Notably, the uni-directional Mamba performs competitively and even slightly surpasses the bi-directional counterpart when trained on 32-frame inputs. On the other hand, the bi-directional model demonstrates stronger performance as video length increases. These results highlight the critical role of the Mamba module and suggest its potential in enabling future designs that support streaming video input for video-LLMs.

\section{Conclusion}
We introduced \method, a novel Video-LLM model that enhances long-video understanding using a Mamba-based temporal encoder and efficient token reduction. By explicitly integrating spatiotemporal dynamics into visual tokens early in the pipeline, \method enables significant token compression while preserving critical information in the compressed inputs. Experiments demonstrate that \method achieves new state-of-the-art results on long-video understanding benchmarks while substantially improving computational efficiency. 

\section{Acknowledgments}
We would like to thank Xin Dong, Matthieu Le, Yunhao Fang, and Dacheng Li for their valuable discussions on this work. Sungjin Ahn was supported by the NRF of Korea funded by the Ministry of Science and ICT (MSIT) (No. 2021H1D3A2A03103645).

{
  \small
  \bibliographystyle{unsrt}
  \bibliography{ref}
}

\clearpage

\section{Appendix}
\subsection{Qualitative Results}
\begin{figure*}[!htbp]
    \centering
    \includegraphics[width=1.0\linewidth]{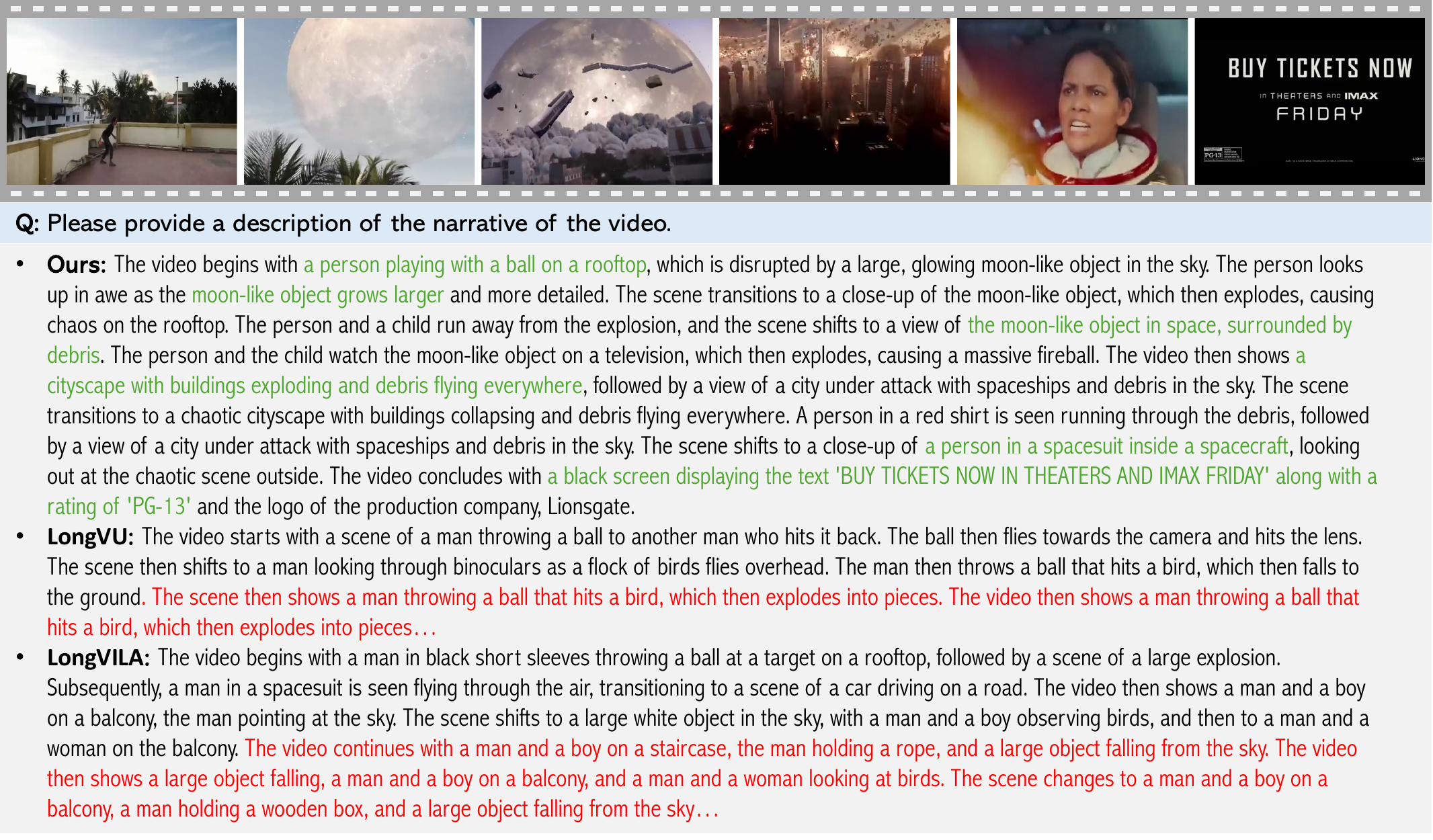}
    \caption{\textbf{Effective Long Video Understanding.} We compare \method + Temporal Sampling with existing long video LLMs. Reults show that \method delivers a more detailed and coherent summary, effectively capturing key events and transitions throughout the film. The example videos can be found in \href{https://research.nvidia.com/labs/lpr/storm}{our website}.}
    \label{fig:appx_qualitative_compare_with_existing}
\end{figure*}

\begin{figure*}[!htbp]
    \centering
    \includegraphics[width=1.0\linewidth]{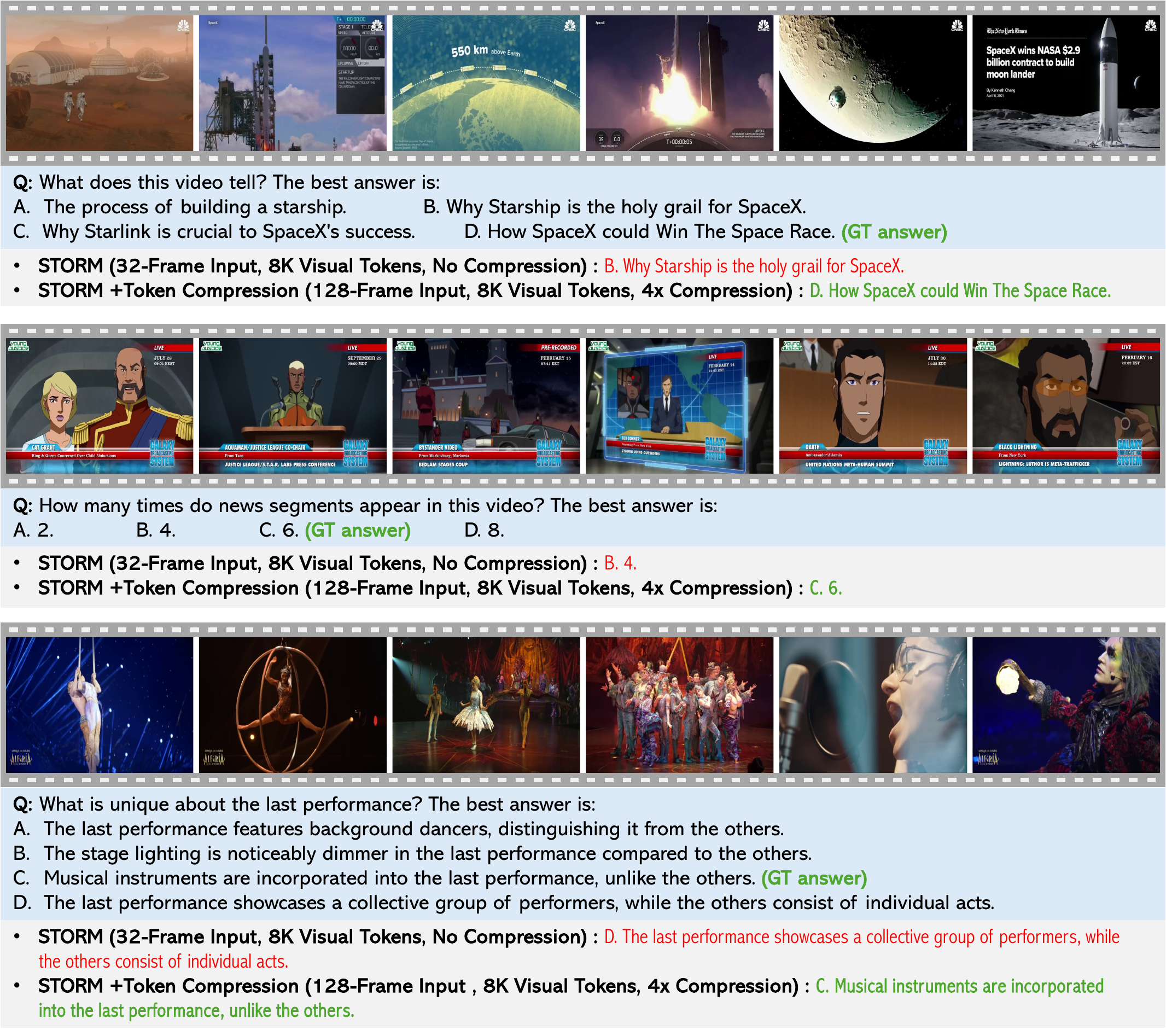}
    \caption{\textbf{Importance of Long Video Context.} We compare \method with a 32-frame input to \method + Temporal Sampling using a 128-frame input. Both configurations have negligible differences in computational cost; however, the latter encodes additional information into compressed tokens due to the extended frame sequence. The examples illustrate that processing more frames allows the model to capture richer temporal dynamics and contextual information. This leads to a stronger understanding of the video's narrative, reduces information loss, and enhances the ability to reason through temporal events across the entire video. The example videos can be found in \href{https://research.nvidia.com/labs/lpr/storm}{our website}.}
    \label{fig:appx_qualitative_num_input_frames}
\end{figure*}

\begin{figure*}[!htbp]
    \centering
    \includegraphics[width=1.0\linewidth]{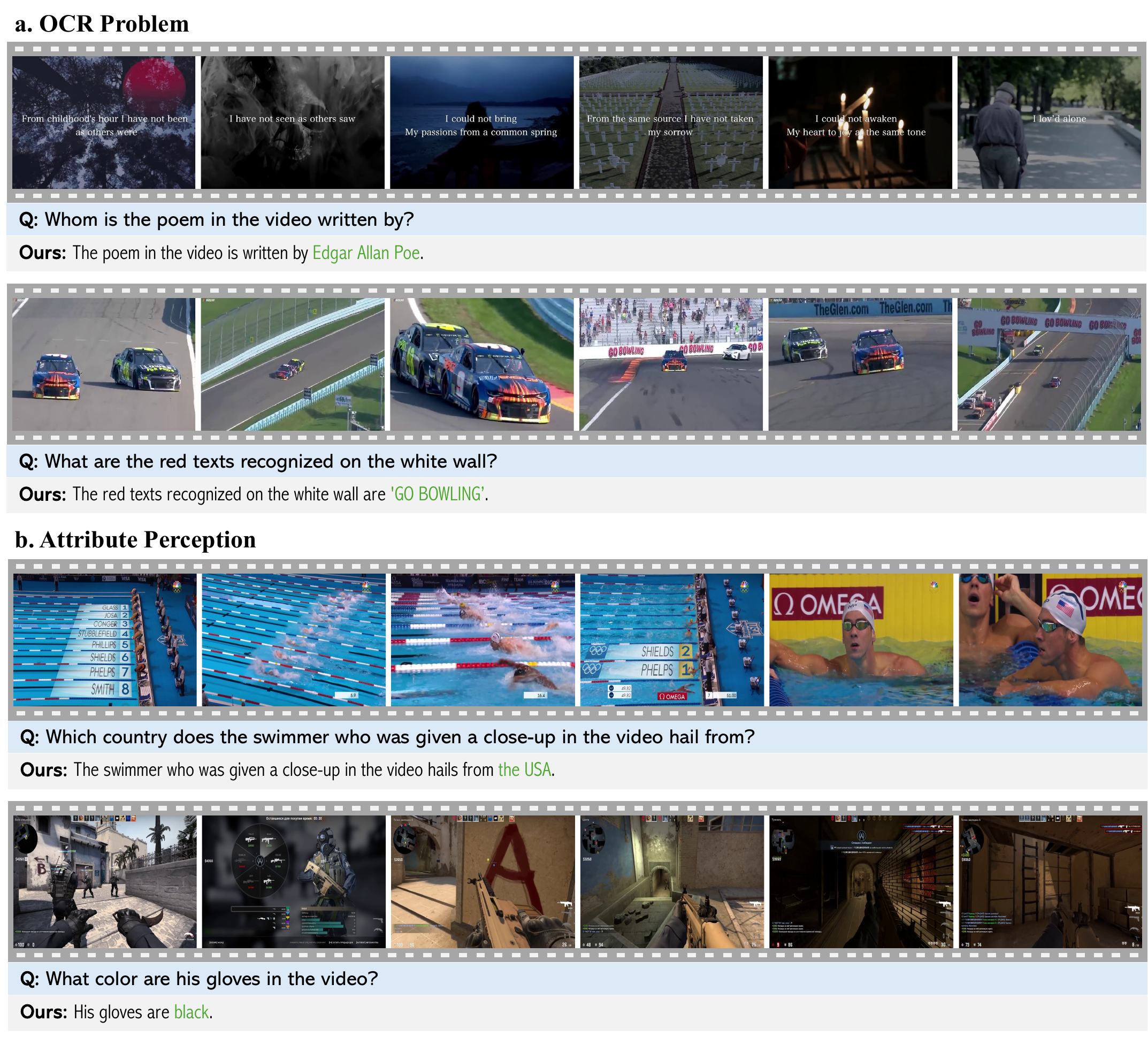}
    \caption{\textbf{Showcase of Video Understanding Abilities in Various Task Categories.} We provide additional examples to showcase model's video understanding capabilities in different aspects. This is done by providing the models with open-ended queries that require the model to generate answers in raw text form without any given choices. Part 1. The example videos can be found in \href{https://research.nvidia.com/labs/lpr/storm}{our website}.}
    \label{fig:appx_qualitative_categories_showcase_1}
\end{figure*}

\begin{figure*}[!htbp]
    \centering
    \includegraphics[width=1.0\linewidth]{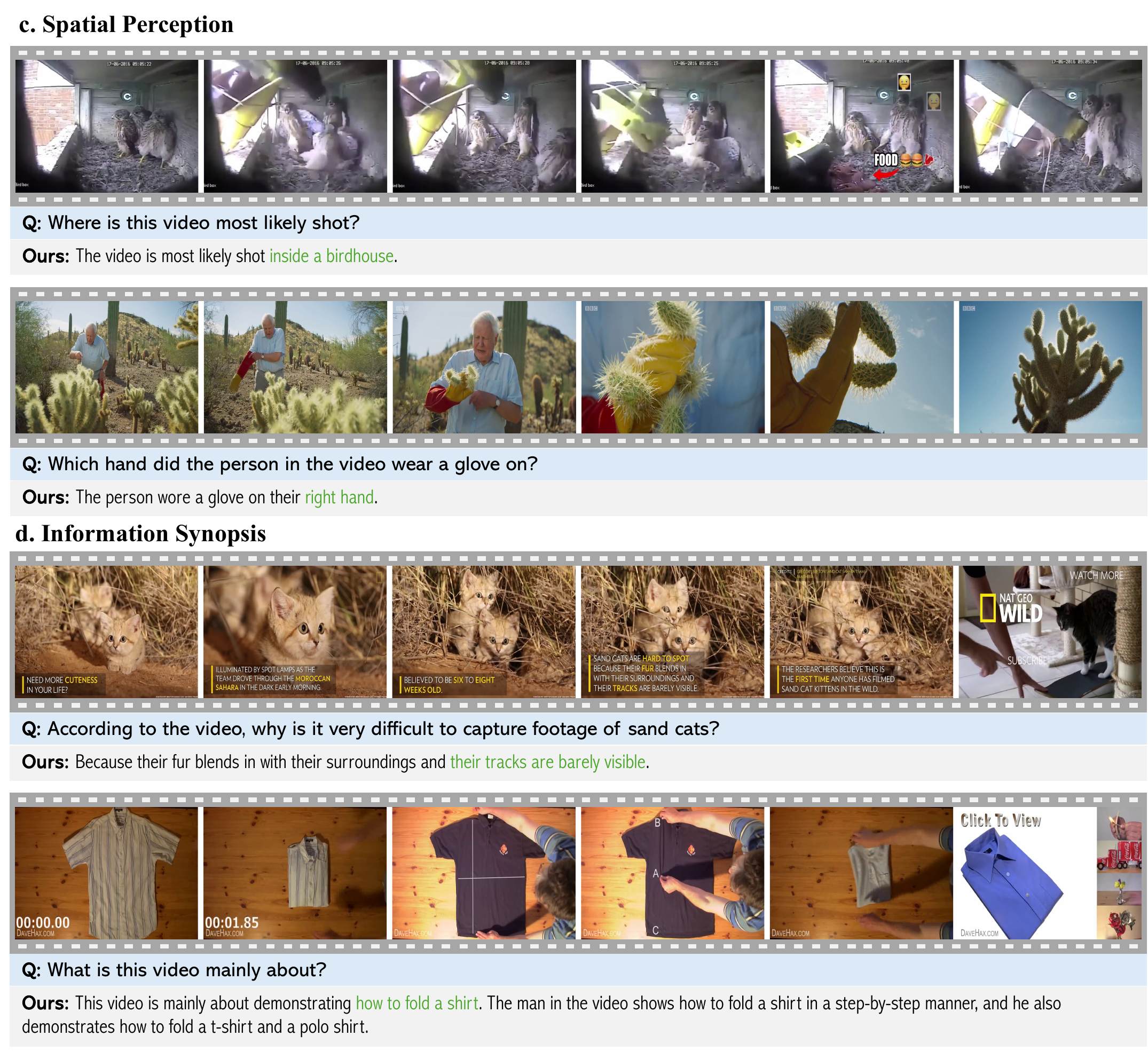}
    \caption{\textbf{Showcase of Video Understanding Abilities in Various Task Categories.} Continue 2. The example videos can be found in \href{https://research.nvidia.com/labs/lpr/storm}{our website}.}
    \label{fig:appx_qualitative_categories_showcase_2}
\end{figure*}

\begin{figure*}[!htbp]
    \centering
    \includegraphics[width=1.0\linewidth]{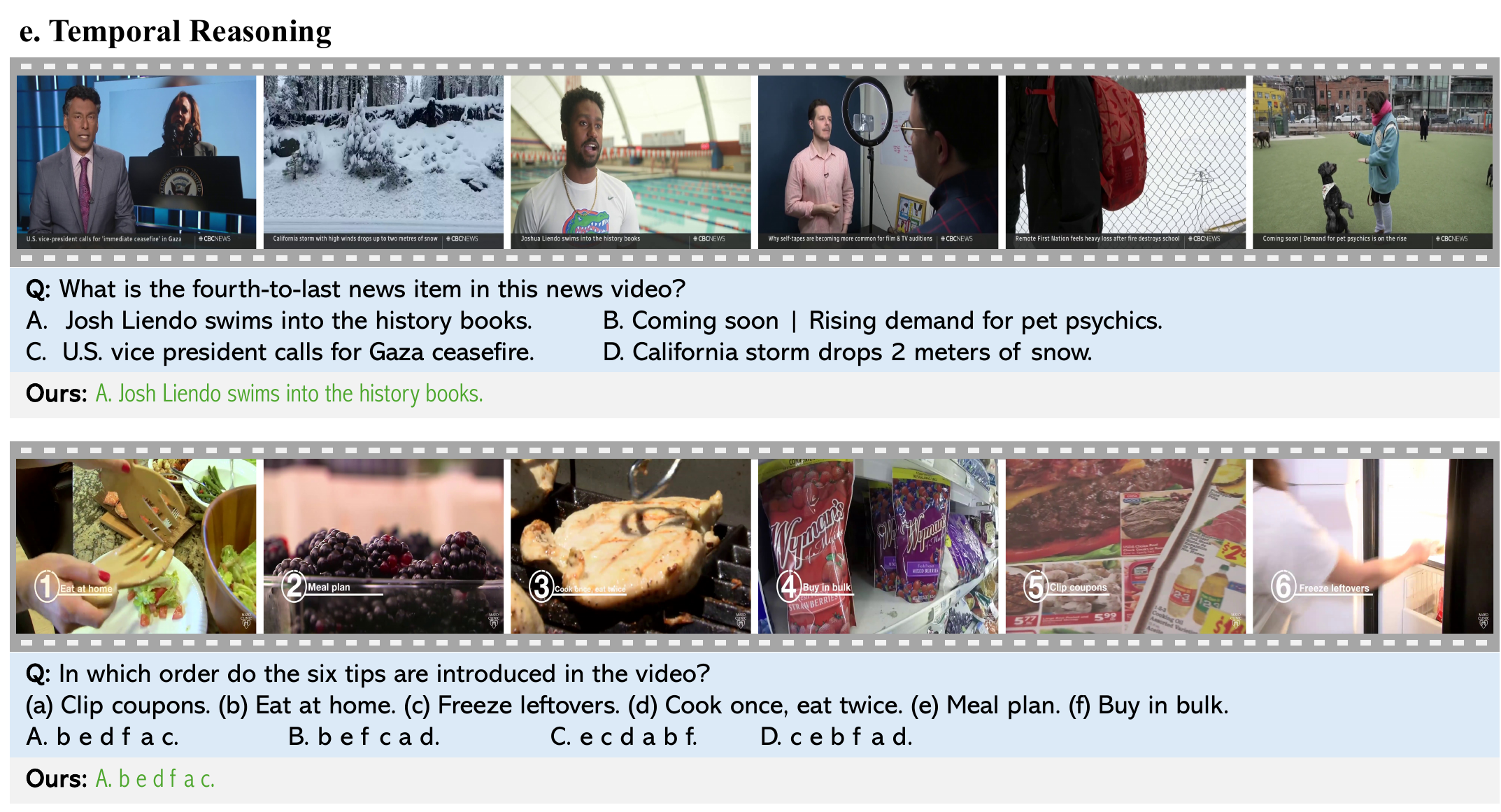}
    \caption{\textbf{Showcase of Video Understanding Abilities in Various Task Categories.} Continue 3. The example videos can be found in \href{https://research.nvidia.com/labs/lpr/storm}{our website}.}
    \label{fig:appx_qualitative_categories_showcase_3}
\end{figure*}

We present comprehensive qualitative evaluations in \autoref{fig:appx_qualitative_compare_with_existing} to~\autoref{fig:appx_qualitative_categories_showcase_3}, which are segmented into three subsections

\begin{enumerate}
    \item \textbf{Effective Long Video Understanding}: Demonstrating \method's ability to effectively utilize long video inputs by comparing it with existing long-video LLMs.
    \item \textbf{Importance of Long Video Context}: Highlighting the need for long video inputs by showcasing scenarios where 128-frame inputs (with token compression) enable accurate predictions, whereas 32-frame inputs fail. 
    \item \textbf{Showcase of Video Understanding Abilities}: Illustrating \method's capabilities in various aspects such as OCR, spatial perception, temporal reasoning, and so on.
\end{enumerate}

\myparagraph{Effective Long Video Understanding.} We compare our proposed \method + Temporal Sampling with LongVILA and LongVU, both designed for long video understanding. We use a short film depicting a "moonfall disaster" from the VILA webpage \footnote{https://vila.mit.edu/}. The models are prompted to provide a narrative description of the video. The short film was chosen for its engaging and dramatic storyline that spans various interconnected scenarios, all contributing to a cohesive narrative. Understanding this video requires the models to comprehend each individual scene and effectively integrate temporal events to grasp the complete story. Both \method and LongVILA use 128 input frames, while LongVU output was obtained from its online demonstration which uses 1fps input.

As shown in \autoref{fig:appx_qualitative_compare_with_existing}, \method delivers the most detailed and coherent summary of the video's narrative, effectively capturing key events and transitions throughout the entire film. Its response showcases a comprehensive understanding of the content, highlighting its ability to connect temporal events across different scenes. In contrast, the baseline models LongVILA and LongVU focus on some of the events but fail to cover all critical moments that contribute to the overall storyline. Their responses also highlight specific scenes without integrating them into the full context. Moreover, we observed that the baseline models often generate redundant content, repeating the same sentences with minimal new information, which reveals their limitations in handling open-ended queries. Notably, our \method with Temporal Sampling is also computationally more efficient. By applying temporal sampling, we reduce the number of tokens to the equivalent of processing 32 frames. This comparison showcases \method's superior ability to leverage long video inputs for in-depth visual understanding.

\myparagraph{Importance of Long Video Context.} 
We further demonstrate the significance of incorporating long video context by providing qualitative examples where a 128-frame input yields more accurate predictions than a 32-frame input, as shown in \autoref{fig:appx_qualitative_num_input_frames}. Using samples from the VideoMME benchmark, we compare two configurations of our \method: one with a 32-frame input without compression, and another with a 128-frame input employing a temporal sampling ratio of 4. In both settings, the number of tokens fed into the LLM remains the same; however, the \method with temporal sampling encodes additional information into the compressed tokens due to the extended frame sequence.

The inclusion of more frames allows the model to capture richer temporal dynamics and contextual information. For example, the 128-frame input enables the model to develop a stronger understanding of the video's narrative (\autoref{fig:appx_qualitative_num_input_frames} top). It also allows the 128-frame model to capture additional events that the 32-frame model misses (\autoref{fig:appx_qualitative_num_input_frames} center). Finally, the additional information further improve model's ability to reason through different temporal events across the entire video to form a coherent understanding (\autoref{fig:appx_qualitative_num_input_frames} bottom). These example demonstrate the crucial role of long video context in tasks that require detailed temporal reasoning and comprehensive content understanding.

\myparagraph{Showcase of Video Understanding Abilities.} 
Finally, we conclude our qualitative evaluation by showcasing the diverse video understanding capabilities of \method, including OCR, attribute perception, spatial perception, information synopsis, and temporal reasoning. Results are shown in \autoref{fig:appx_qualitative_categories_showcase_1} to~\autoref{fig:appx_qualitative_categories_showcase_3}. We use the same setting of \method + Temporal Sampling with 128-frame input and sampling ratio of 4. Utilizing videos from the VideoMME benchmark, we designed a more challenging assessment to thoroughly evaluate the model's proficiency. Instead of providing the model with multiple-choice questions accompanied by predefined answer options, we transformed these tasks into open-ended queries that require the model to generate answers in raw text form without any given choices. This modification significantly increases the task's difficulty, as it demands a precise understanding of the content and the ability to accurately locate and extract specific information from the video input.

Our qualitative results demonstrate that \method provides strong performance in these scenarios. Despite the increased complexity, the model effectively interprets intricate visual details, recognizes textual information within videos, and provides coherent summaries of temporal events. This showcases \method's robust ability to handle various aspects of video understanding.

\subsection{Additional Results} \label{sec:appx_dataset_composition}

\myparagraph{Ablation on Token Budget and Token Compression Strategies.} \autoref{tab:main_results_full} extends \autoref{tab:comp_abl} in the main text by providing a comprehensive comparison of different compression method combinations across various token budgets during training. Overall, considering both compression ratio and inference latency, we find that \method with temporal pooling (\method + T. Pooling) is the most efficient and effective approach. Additionally, test-time temporal sampling offers a lossless way to further enhance inference efficiency in inference time.

\begin{figure*}[!htbp]
    \centering
    \includegraphics[width=\textwidth]{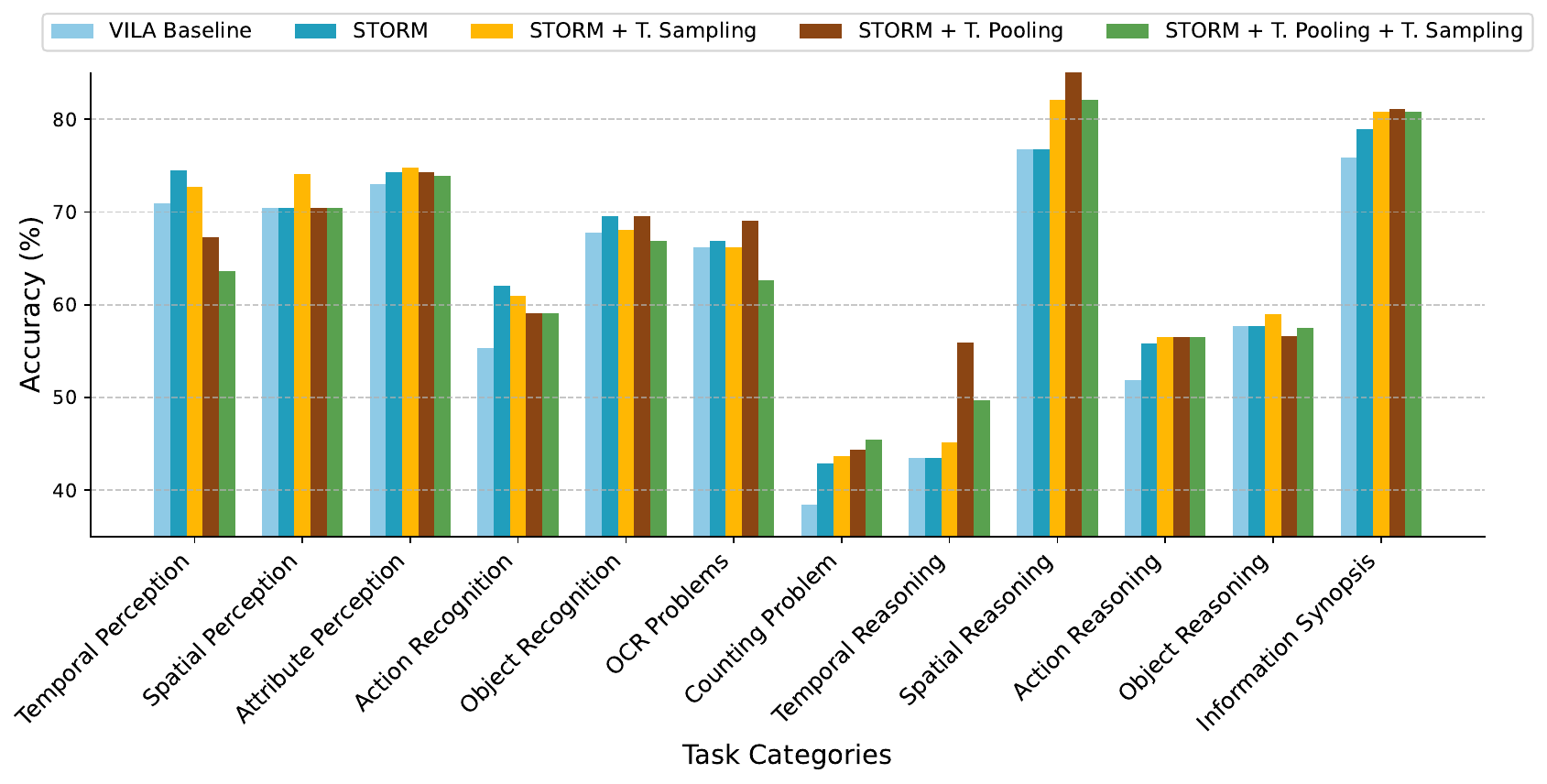}
    \caption{\textbf{VideoMME Results by Task Categories.}}
    \label{fig:videomme_category}
\end{figure*}

\myparagraph{Task-Level Analysis on VideoMME} \autoref{tab:results_length} shows the VideoMME results with different video lengths. The short is less than 2 minutes, the medium is up to 15 minutes, and the long is up to 60 minutes. Overall, our \method with token compression outperforms the VILA baseline and \method with no token compressions for all video lengths. 
\autoref{fig:videomme_category} compares the VideoMME results by task categories. We find that \method with temporal pooling especially improves the object reasoning task accuracy, and \method with test-time temporal sampling improves the attribute perception accuracy. Both token compression methods improve the temporal perception task accuracies compared to VILA and \method. It indicates that the temporal perception task requires a longer video context, and our token compression methods are effective for such tasks. 

\myparagraph{Effect of Dataset Composition} \autoref{tab:dataset_effect} shows how dataset composition affects model performance during 128 frames fine-tuning. We compare using the full LLaVA-Video dataset ($\sim$~1.35M samples) versus only its longest 25\% videos with at least 128 frames ($\sim$~360K samples). Interestingly, while \method improves with the larger dataset across all benchmarks, the baseline model actually performs worse on several benchmarks when trained on the full dataset.

Two key differences between these datasets are size and video length distribution, where the full dataset contains more data but with a mixture of short and long videos, whereas the long-video subset exclusively consists of longer videos. Since larger, more diverse datasets typically improve performance (assuming similar data quality), we attribute the baseline model's unexpected performance drop to its limited ability to generalize from shorter to longer videos. More specifically, when trained predominantly on shorter clips from the full dataset, the baseline overfits and can not effectively handle long contexts at inference time. Training solely on longer videos dataset variant better aligns with test conditions, partially addressing this limitation.

In contrast, \method shows consistent performance gains in all benchmarks when trained on the larger and more diverse data set. This suggests that \method is more robust in handling longer sequences and is capable of using a wide range of video lengths to enhance its overall performance.

\subsection{Architecture Details}

\begin{figure*}[!htbp]
    \centering
    \includegraphics[width=0.85\linewidth]{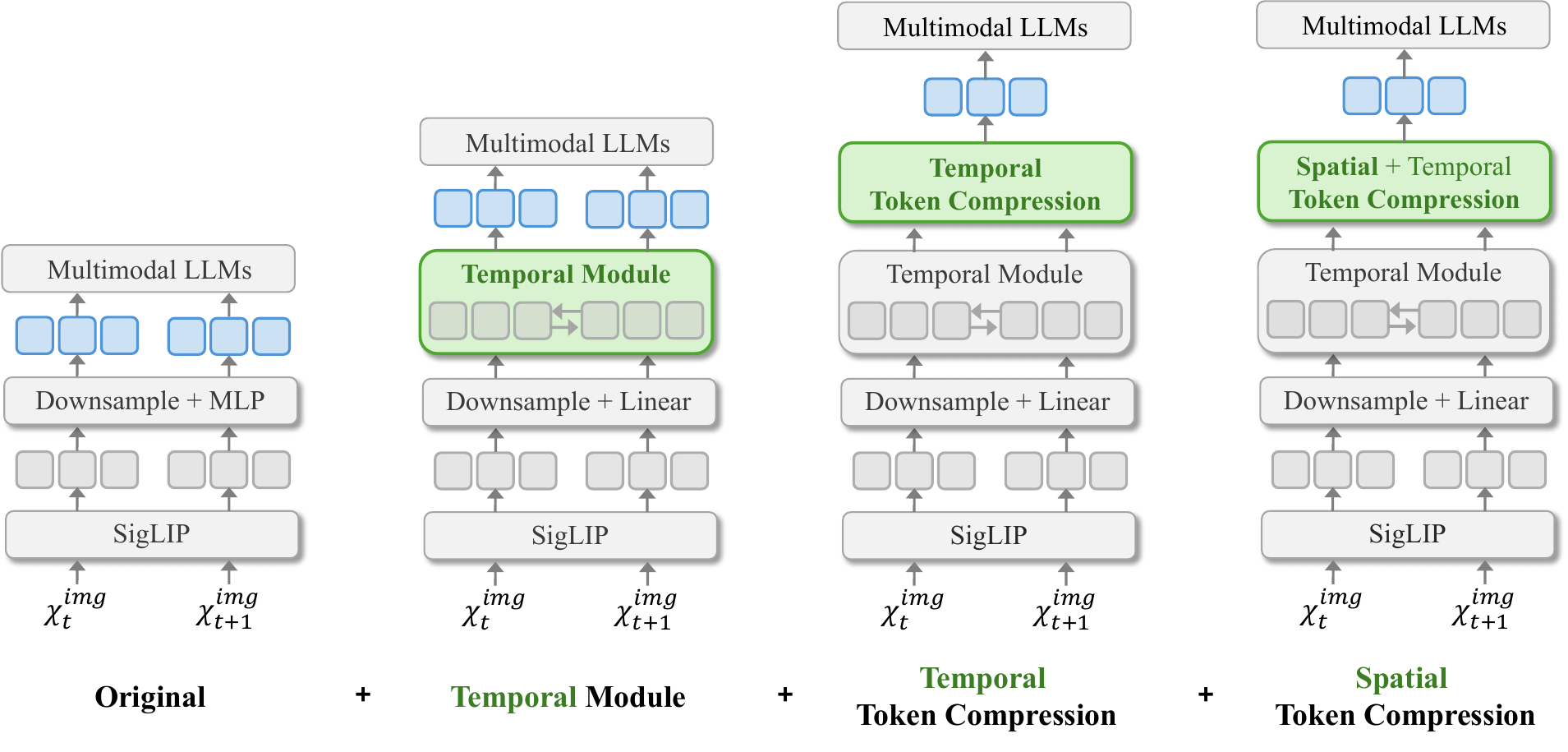}
    \caption{ \textbf{Breaking Down the \method Architecture.} We begin with a standard multimodal pipeline that uses a pixel-shuffle downsampling layer and an MLP projector. In \method, we replace the MLP with a linear layer and introduce our Mamba-based temporal module on top. Since the Mamba layer propagates spatiotemporal information in each visual tokens, the model can then perform temporal and spatial token compression of these tokens before passing them to the LLM, allowing \method to handle longer sequences more efficiently.}
    \label{fig:appx_architecture}
\end{figure*}

\method is built on a standard multimodal pipeline but introduces key modifications for improved reasoning ability and token efficiency. \autoref{fig:appx_architecture} illustrates the detailed composition of our models. Instead of an MLP projector, \method uses a linear layer followed by a Mamba-based temporal module projector which integrates spatiotemporal information into visual tokens.  

\method  incorporates three main components: (1) The Mamba-Based Temporal Projector captures and propagates spatiotemporal information within visual tokens. (2) \textbf{Temporal Token Compression Module} applies compression on temporal dimension using training-based average pooling and/or training-free sampling (applied only at test time). (3) \textbf{Spatial Token Compression} further reduces token number by performing training-based frame-level spatial average pooling. Both spatial and temporal compression methods—whether training-free or training-based—are independently applicable. Notably, spatial and temporal pooling can be applied in parallel after the Mamba module, while temporal sampling is performed separately at test time. These components enable \method to process longer sequences more efficiently before passing them to the LLM.

\subsection{Datasets Information} \label{sec:appx_dataset}
For all models, we begin with an alignment stage to align the multi-modal projector using the LLaVA-CC3M-Pretrain-595K dataset~\citep{liu2024visual}. Following this, we proceed to the visual instruction fine-tuning stage, experimenting with two different training data mixtures. These SFT mixtures incorporate both image and video data, encompassing three task types: captioning, open-ended question answering, and multiple-choice question answering. Further details are provided in the following:

\begin{itemize}
    
    \item \textbf{SFT Data:} For most of our main experiments, we construct an expanded mixture by incorporating additional high-quality image datasets—such as Cambrian-1375K~\citep{tong2024cambrian1}, Idefics2-SFT~\citep{laurençon2024idefics2}, and LLaVA-OneVision-Images-SFT~\citep{li2024llavaonevision}—along with video datasets including M4-Instruct-Video~\citep{zhang2024direct} and Youtube~\citep{zhu2023languagebind}. This enlarged dataset is used to scale up training and enhance overall performance. Detailed compositions of these mixtures are provided in \autoref{tab:appendix-data}.

    \item \textbf{Long Video SFT Data:} In the long video fine-tuning stage, our goal is to adapt models initially trained on full SFT data with 32-frame inputs to handle 128-frame inputs. Because processing 128-frame input incurs significant computational cost, we reduce training time by using a smaller dataset at this stage. Specifically, we select videos from only the
    LLaVA-Video dataset~\citep{zhang2024llavavideo} that contains data amount roughly 11\% of the full dataset (approximately 1.35M video-text pairs).

    \item \textbf{Long Video 25\% Data:} As an ablation study to investigate the impact of dataset composition in the long video fine-tuning stage, we introduce an additional dataset derived from the LLaVA-Video dataset~\citep{zhang2024llavavideo}. This subset consists exclusively of long videos with at least 128 frames, comprising approximately 25\% of the full dataset (around 360K video-text pairs). Unlike the Long Video SFT Data, which includes both short and long videos, this dataset contains only long videos. Our experiments in Section~\ref{sec:appx_dataset_composition} and \autoref{tab:dataset_effect} reveal distinct behaviors in our models and baselines in the composition of the dataset.
\end{itemize}

\begin{table*}[h]
\setlength{\tabcolsep}{3pt}
\small\centering
\begin{tabular}{c}
\toprule
Datasets \\ \midrule
\makecell{LLaVA-SFT~\citep{liu2024visual}, Idefics2-SFT~\cite{laurençon2024idefics2}} \\ 
\makecell{MSR-VTT~\cite{xu2016msr}, Image Paragraph Captioning~\cite{krause2017hierarchical}, ShareGPT4V-100K~\cite{chen2023sharegpt4v}} \\ 
\makecell{CLEVR~\cite{johnson2017clevr}, NLVR, VisualMRC~\cite{tanaka2021visualmrc}} \\ 
\makecell{ActivityNet-QA~\cite{yu2019activitynet}, LLaVA-OneVision-Images-SFT~\cite{li2024llavaonevision}, \\ iVQA~\cite{yang2021just}, MSRVTT-QA, STEM-QA~\cite{shen2024measuringvisionlanguagestemskills}} \\ 
\makecell{DVQA~\cite{kafle2018dvqa}, ST-VQA~\cite{biten2019scene}, SynthDoG-en~\cite{kim2022synthdog-en}, TextOCR-GPT4V~\cite{textocr-gpt4v}, MTWI} \\ 
\makecell{ScienceQA-train~\cite{lu2022scienceqa}, VQAv2-train, ViQuAE~\cite{lerner2022viquae}, Visual Dialog~\cite{das2017visualdialog}, \\ GQA-train~\cite{hudson2019gqa}, ChatQA~\cite{masry2022chartqa}, Geo170K~\cite{gao2023gllavasolvinggeometricproblem}, \\ LRV-Instruction~\cite{liu2024mitigatinghallucinationlargemultimodal}, RefCOCO-train~\cite{yu2016modelingcontextreferringexpressions}, \\ DocVQA~\cite{mathew2021docvqa}, GeoQA~\cite{chen-etal-2021-geoqa}, KVQA~\cite{marino2019okvqa}, Cambrian-1375K~\cite{tong2024cambrian1}} \\ 
\makecell{AI2D~\cite{Kembhavi2016ai2d}, Shikra~\cite{chen2023shikra}, Unimm-Chat~\cite{yu2023UniMM-Chat}} \\ 
\makecell{LRV-Instruction~\cite{liu2023lrv}, SVIT~\cite{zhao2023svit}, MMC-Instruction~\cite{liu2024mmcadvancingmultimodalchart}, \\ M4-Instruct-Images~\citep{liu2024llavanext}, M4-Instruct-Video~\citep{zhang2024direct} , WIT~\cite{Srinivasan_2021}, Youtube~\citep{zhu2023languagebind}, etc} \\
\bottomrule
\end{tabular}
\caption{SFT data mixture.}
\label{tab:appendix-data}
\end{table*}

\begin{table*}[h]
\setlength{\tabcolsep}{1.3pt}
\footnotesize\centering
\begin{adjustbox}{width=0.95\textwidth}
\begin{tabular}{lcccccccccccc}
\toprule
\multirow{2}{*}{\textbf{Models}} & \textbf{Size} & \textbf{Comp.} & {\textbf{Latency}} & \textbf{\#Frames }  & \textbf{\# Frames} & \textbf{MVBench} & \textbf{MLVU} & \textbf{LongVideoBench} & \textbf{VideoMME} \\
\cmidrule(lr){7-10}
  &   & \textbf{Ratio (\%) } & \textbf{(s)} & \textbf{(train)} & \textbf{(test)} & \textbf{test}  & \textbf{dev} & \textbf{val} & \textbf{(w/o sub.)} \\
\midrule
\textbf{Duration} & &  &  & &  & 16 sec & 3$\sim$120 min & 8 sec$\sim$60 min & 1$\sim$60 min \\
\midrule
\textbf{Token Budget: 8K } \\
VILA Baseline  & 7B & 100 & 4.31 & 32 & 256 & 69.5 & 70.2 & 55.9 & 60.1 \\
\rc \method  & 7B & 100 & 4.47 & 32 &  256  & 70.3 & {71.1} & 54.5 & 62.5 \\
\rc \method+ S. Pooling & 7B & 25 & 1.82 & 128 & 256  & 63.9 & 67.9 & 54.5 & 57.5 \\
\rc \method+ T. Pooling & 7B & 25 & 1.82 & 128 & 256  & \textbf{71.3} & {72.5} & 59.5 & \textbf{63.4} \\
\rt \method  + T. Sampling *  & 7B & 50 & 2.50 & 32 & 256 & 70.1 & 70.8 & 54.8 & \textbf{63.1} \\
\rt \method+ S. Pooling + T. Sampling * & 7B & 12.5 & 1.51 & 128 & 256 & 65.2 & 68.3 & 55.0 & 57.6 \\
\rt \method+ T. Pooling + T. Sampling * & 7B & 12.5 & 1.51 & 128 & 256 & 70.6 & \textbf{72.9} & \textbf{60.5} & {62.4} \\
\midrule
\textbf{Token budget: 2K} \\
\rc \method + S. Pooling  & 7B & 25 & 1.82 & 32 & 256 & {68.9} & {69.2} & 56.0 & {61.1} \\
\rc \method + T. Pooling   & 7B & 25 & 1.82 & 32 &  256 & \textbf{70.4} & {71.0} & 54.2 & {61.2} \\
\rt \method + S. Pooling + Sampling * & 7B & 12.5 & 1.51 & 32  & 256 & 68.9 & 69.5 & {56.3} & 60.9 \\
\rt \method + T. Pooling + Sampling * & 7B  & 12.5  & 1.51 & 32 & 256 & 68.9 & 69.5 & {56.3} & 60.9 \\
\midrule
\textbf{Token budget: 0.5K} \\
\rc \method + S. Pooling + T. Pooling & 7B & 6.25 & 1.36 & 32  & 256 & 68.5 & 68.2 & 53.7 & 60.2 \\
\bottomrule
\multicolumn{10}{l}{\small * 2x additional compression at test time.} \\
\end{tabular}
\end{adjustbox}
\caption{\textbf{Ablation on Token Budget and Token Compression Strategies.} Both spatial and temporal pooling are with $4 \times$ compression. The number of frames used during testing is consistent across models but can differ across tasks. The \# frames is the maximum number of frames during testing. We summarize the details of the number of frames for each task in \autoref{tab:num_frames_summary}. The temporal token sampling is with $2 \times$ additional compression. } 
\label{tab:main_results_full}
\end{table*}

\begin{table}[!htbp]
    \centering
    \footnotesize
    \setlength{\tabcolsep}{2pt}
    \begin{adjustbox}{width=0.95\linewidth}
    \begin{tabular}{lccc}
        \toprule
        \textbf{Models} & \textbf{8F} & \textbf{32F (T. Pooling)} & \textbf{128F (T. Pooling)} \\
        \midrule
        \textbf{MVBench} \\
        VILA (w/o Mamba) & 67.9 & 68.7  & 68.1  \\
        \method (w/ Mamba) & \textbf{68.8} & \textbf{70.4}  & \textbf{71.3}  \\
        \midrule
        \textbf{MLVU} \\
        VILA (w/o Mamba) & \textbf{67.7} & \textbf{71.0}  & 69.9  \\
        \method (w/ Mamba) & 66.8 & \textbf{71.0}  & \textbf{72.5}  \\
        \midrule
        \textbf{LongVidBench} \\
        VILA (w/o Mamba) & \textbf{52.4} & \textbf{55.4}  & 57.4  \\
        \method (w/ Mamba) & 50.6 & 54.2  & \textbf{59.5}  \\
        \midrule
        \textbf{VideoMME} \\
        VILA (w/o Mamba) & 60.0 & 58.9  & 61.7  \\
        \method (w/ Mamba) & \textbf{60.2} & \textbf{61.2}  & \textbf{63.4}  \\
        \midrule
        \textbf{Avg} \\
        VILA (w/o Mamba) & \textbf{62.0} & 63.5  & 64.3  \\
        \method (w/ Mamba) & 61.6 & \textbf{64.2}  & \textbf{66.7}  \\
        \bottomrule
    \end{tabular}
    \end{adjustbox}
    \caption{\textbf{Detailed Comparison across Benchmarks for \autoref{tab:mamba_compression}.} \method consistently improves performance across all benchmarks as the input video length increases from 8F to 32F to 128F. In contrast, the baseline VILA exhibits diminishing gains with longer inputs and even experiences performance degradation on certain benchmarks when extending from 32F to 128F. These results highlight the critical role of the Mamba module in effectively leveraging long-video inputs to enhance model performance.}

    \label{tab:mamba_compression_full}
\end{table}

\begin{table*}[!htbp]
\footnotesize\centering
\begin{adjustbox}{width=\textwidth}
\begin{tabular}{lcllll}

\toprule
\multirow{2}{*}{\textbf{Models}} & \textbf{Dataset} & \textbf{MVBench} & \textbf{MLVU} & \textbf{LongVideoBench} & \textbf{VideoMME} \\
\cmidrule(lr){3-6}
  & \textbf{type} & \textbf{test}  & \textbf{dev} & \textbf{val} & \textbf{(w/o sub.)} \\
\midrule
\textbf{Duration} & & 16 sec & 3$\sim$120 min & 8 sec$\sim$60 min & 1$\sim$60 min \\
\midrule
VILA Baseline + T. Pooling  & long-video only (25\% of full) & {67.2} & {71.4} & {59.2} & {62.2} \\
VILA Baseline + T. Pooling  & full LLaVA-Video~\citep{zhang2024llavavideo} & 68.1 (\textcolor{blue}{+0.9}) & 69.9 (\textcolor{red}{-1.5}) & 57.4 (\textcolor{red}{-1.8}) & {61.7} (\textcolor{red}{-0.5}) \\
\midrule
VILA Baseline + T. Pooling + T. Sampling * & long-video only (25\% of full)  & 64.5 & 71.4 & 59.2 & {61.0} \\
VILA Baseline + T. Pooling + T. Sampling *  & full LLaVA-Video~\citep{zhang2024llavavideo} & 67.8 (\textcolor{blue}{+3.3}) & 70.1 (\textcolor{red}{-1.3}) & 57.7 (\textcolor{red}{-1.5}) & 59.3 (\textcolor{red}{-1.7}) \\
\midrule
\method+ T. Pooling  & long-video only (25\% of full) & {69.4} & {71.7} & {57.6} & {63.2} \\
\method+ T. Pooling  & full LLaVA-Video~\citep{zhang2024llavavideo} & \textbf{71.3} (\textcolor{blue}{+1.9}) & {72.5} (\textcolor{blue}{+0.8}) & {59.5} (\textcolor{blue}{+1.9}) & \textbf{63.4} (\textcolor{blue}{+0.2}) \\
\midrule
\method+ T. Pooling + T. Sampling * & long-video only (25\% of full)  & 68.8 & {72.7} & {59.2} & {62.6} \\
\method+ T. Pooling + T. Sampling *  & full LLaVA-Video~\citep{zhang2024llavavideo} & 70.6 (\textcolor{blue}{+1.8}) & \textbf{72.9} (\textcolor{blue}{+0.2}) & \textbf{60.1} (\textcolor{blue}{+0.9}) & {62.4} (\textcolor{red}{-0.2}) \\
\bottomrule
\multicolumn{6}{l}{\small * 2x additional compression at test time.} \\
\end{tabular}
\end{adjustbox}
\caption{\textbf{Effect of Dataset Composition on 128-frame Fine-Tuning.} We compare models trained on two variants: the ``full LLaVA-Video'' dataset~\citep{zhang2024llavavideo} ($\sim$1.35M video-text pairs) versus the ``long-video only'' subset using top 25\% longest videos (minimum of 128 frames) ($\sim$360K pairs). Values in parentheses show performance differences between training on the long-video subset vs the full dataset. \method consistently benefits from the larger, more diverse dataset across benchmarks, while the baseline VILA model degrades on several benchmarks when trained on the full dataset.}
\label{tab:dataset_effect}
\end{table*}

\begin{table*}[!htbp]
\setlength{\tabcolsep}{2pt}
\small \centering
\begin{adjustbox}{width=0.7\linewidth}
\begin{tabular}{lccccc}
\toprule
\textbf{Models} & \textbf{\# frames } & \textbf{Short}   & \textbf{Medium}  & \textbf{Long} & \textbf{Avg.}\\
& \textbf{(train)} & $<$ 2 min & 4$\sim$15 min & 30$\sim$60 min & \\
\midrule
\textbf{Token Budget: 8K } \\
VILA Baseline & 32  & 73.0 & {58.0} & 49.2 & 60.1 \\
\method & 32   & \textbf{75.6} & {60.9} & 51.1 & 62.5\\
\method + T. Sampling* & 32  & 75.2 & {60.8} & 53.2 & {63.1} \\
\method + T. Pooling & 128  & 72.4 & \textbf{64.4}  & \textbf{53.4} & \textbf{63.4} \\
\method + T. Pooling + T. Sampling* & 128 & {72.9} & {60.9} & \textbf{53.4} & {62.4}  \\
\bottomrule
\multicolumn{6}{l}{\small * 2x additional compression at test time.} \\
\end{tabular}
\end{adjustbox}
\caption{\textbf{Breakdown of VideoMME Results by Input Video Length.}
} 
\label{tab:results_length}
\end{table*}

\begin{table*}[!htbp]
\footnotesize
\centering
\begin{tabular}{cccccc}
\hline
\textbf{\# Frames} & \textbf{Compression Ratio} & \textbf{Overall (ms)} & \textbf{llm (ms)} & \textbf{vision\_tower (ms)} & \textbf{mm\_projector (ms)} \\
\hline
4                & 1                         & 162.92               & 103.80            & 52.41                      & 6.71 \\
8                & 1                         & 270.87               & 174.61            & 85.11                      & 11.15 \\
16               & 1                         & 486.37               & 321.73            & 144.47                     & 20.17 \\
32               & 1                         & 933.99               & 623.41            & 269.49                     & 41.09 \\
64               & 1                         & 1910              & 1310           & 515.17                     & 82.41 \\
128              & 1                         & 4270              & 3090           & 1020                    & 163.34 \\
256              & 1                         & 10340             & 7960           & 2030                    & 348.22 \\
512              & 1                         & 28620             & 23710          & 4090                    & 811.31 \\
\midrule
32                & 4                         & 486.97               & 175.75            & 269.96                     & 41.26 \\
64               & 4                         & 920.10               & 322.22            & 515.82                     & 82.06 \\
128               & 4                         & 1800              & 622.29            & 1010                    & 163.23 \\
256               & 4                         & 3680              & 1310           & 2020                    & 348.52 \\
512              & 4                         & 7950              & 3080           & 4060                    & 811.84 \\
\midrule
64               & 8                         & 772.18               & 175.27            & 514.59                     & 82.32 \\
128              & 8                         & 1500              & 322.73            & 1020                    & 163.09 \\
256              & 8                         & 3000              & 622.56            & 2030                    & 348.71 \\
512              & 8                         & 6200              & 1310           & 4070                    & 815.08 \\
\hline
\end{tabular}
\caption{\textbf{Full Latencies on Various Compression Ratios and Input Frames.}}
\label{tab:latencies_compression}
\end{table*}

\begin{figure}[!htbp]
    \centering
    \includegraphics[width=.46\textwidth]{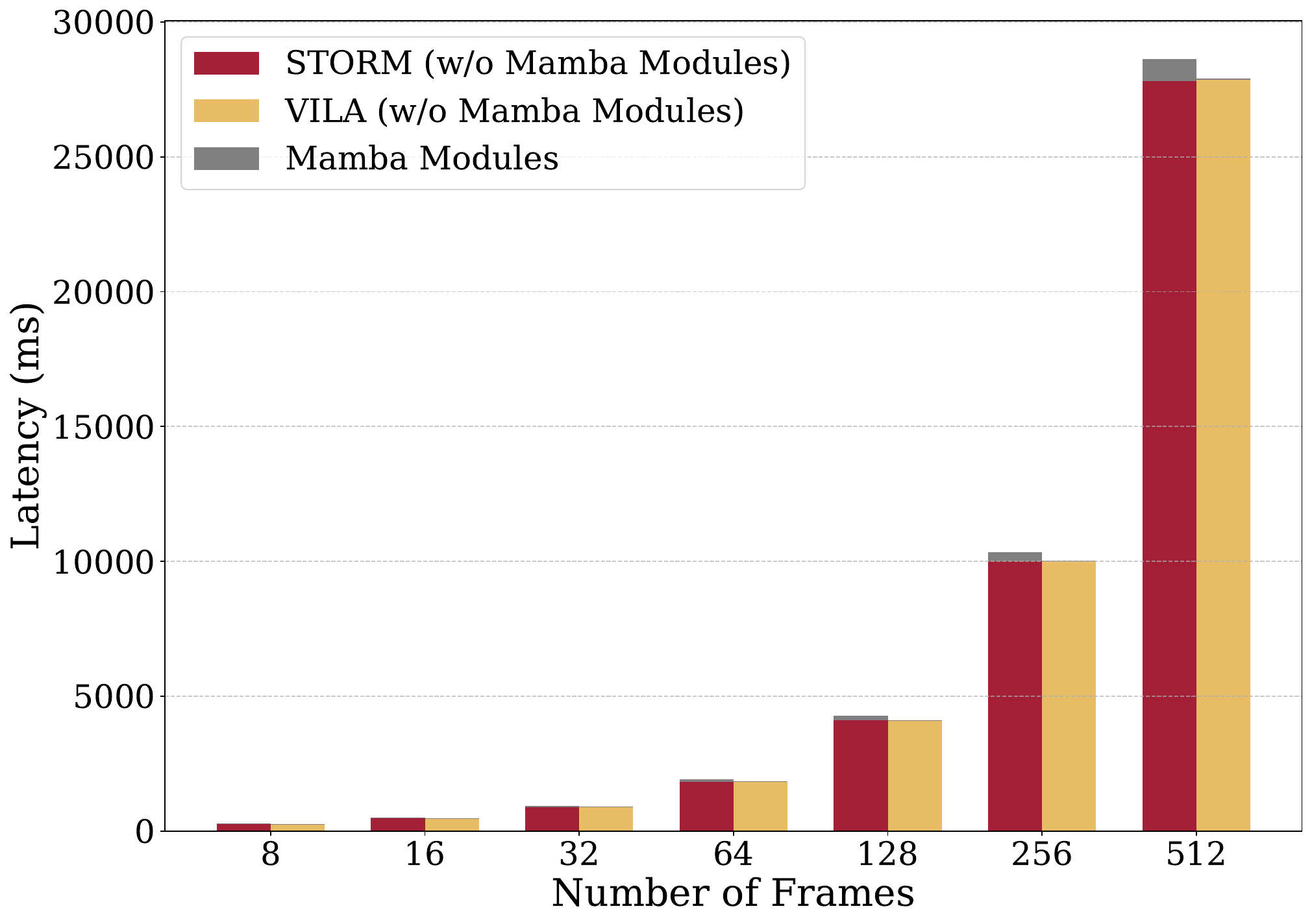}
    \caption{Latency Comparison: VILA vs \method. The multi-modal projector in VILA is a 2-layer MLP, while it is the Mamba Temporal Module in \method.}
    \label{fig:mamba_overhead}
\end{figure}

\subsection{Mamba Temporal Module Latency}
In this section, we compare the latencies of the vanilla VILA architecture and \method across varying numbers of frames without token compression and provide a breakdown of the percentage contribution of the multi-modal projector. All experiments are conducted on a single NVIDIA DGX A100-80G. The results, shown in~\fig{fig:mamba_overhead}, demonstrate that \method incurs negligible overhead compared to the vanilla VILA architecture, with the introduced Mamba Temporal Module accounting for no more than 3\% of the total latency.

\subsection{Inference Details}
\autoref{tab:num_frames_summary} summarizes the number of frames used for inference. We evaluate all models between 8 and 512 frames and select the number of frames with the best accuracy overall for each task and setup. 

\subsection{\method vs Other Temporal Fusion Strategies.} 
In this section, we present early explorations of temporal fusion strategies and show that the simple Temporal-Pooling design used in our final model is surprisingly effective when combined with the Mamba module. Specifically, we experimented with more complex fusion approaches, including TSM~\citep{lin2019tsm} and SlowFast~\citep{feichtenhofer2018slowfast}, using LLaMA3-8B + SigLip. TSM incorporates temporal information by shifting tokens from neighboring frames into the image-based visual encoder. On the other hand, SlowFast encodes video using two token streams: one with high temporal but low spatial resolution, and the other with the opposite configuration.

The results of these experiments are shown in \autoref{tab:fusion_strategies}. These evaluations were conducted during the early stages of our study, and as such, the number of input frames was not consistently matched across variants, making exact comparisons difficult. However, the settings are generally favorable to the TSM and SlowFast variants, as they use the same or more frames than the baseline, which does not perform any temporal fusion. Despite this, both fusion methods fail to yield meaningful improvements over the VILA baseline. In contrast, our STORM design achieves a significant gain, outperforming all other variants.

\begin{table}[!htbp]
\centering
\scriptsize
\caption{Comparison of temporal fusion strategies.}
\label{tab:fusion_strategies}
\resizebox{1\linewidth}{!}{
\begin{tabular}{ccc|cc}
\toprule
\multicolumn{3}{c|}{\textbf{16 frames}} & \multicolumn{2}{c}{\textbf{32-64 frames}} \\
\midrule
\multirow{2}{*}{\textbf{Baseline}} & \textbf{Baseline} & \textbf{TSM} & \textbf{SlowFast} & \textbf{STORM} \\
\textbf{} & \textbf{(T.pool) } & \textbf{(T.pool)} & \textbf{(T.pool)} & \textbf{(T.pool)} \\
\midrule
52.0 & 50.0  & 49.0  & 51.3  &  \textbf{56.8} \\
\bottomrule
\end{tabular}
}
\end{table}

\begin{table*}[!htbp]
\setlength{\tabcolsep}{2pt}
\footnotesize\centering
\begin{tabular}{lcccc}
\toprule
\textbf{Models}  & \textbf{MVBench} & \textbf{MLVU} & \textbf{LongVidBench} & \textbf{VideoMME} \\
\midrule
8-frame-models  & 8 & 64 & 32 & 64 \\
+ Temporal Sampling & 16  & 256  & 256  & 128 \\
\midrule
32-frame-models   & 16 & 64 & 64 & 64 \\
+ Temporal Sampling  & 32 & 256 & 256 & 128 \\
\midrule
32-frame-models + Temporal Pooling  & 32 & 64 & 128  & 64 \\
+ Temporal Sampling  & 64  & 256 & 256 & 128 \\
\bottomrule
\end{tabular}
\caption{\textbf{The Number of Frames Used for Inference.} We evaluate all models for [8, 16, 32, 64, 128, 256, 512] frames and select the best overall for each task and setup.}
\label{tab:num_frames_summary}
\end{table*}

\end{document}